\documentclass[journal]{IEEEtran}
\usepackage{graphicx}
\usepackage{bm}
\usepackage[svgnames,dvipsnames,table]{xcolor}
\usepackage{hyperref}
\usepackage{amsmath}
\usepackage{amssymb}
\usepackage[numbers]{natbib}
\usepackage{adjustbox}
\usepackage{multirow}
\usepackage{tikz}
\usepackage{pgfplots}
\usepgfplotslibrary{patchplots}
\usetikzlibrary{trees,shapes,snakes,calc,matrix,positioning,fit,arrows,patterns,backgrounds,quotes,arrows.meta}
\usepackage{siunitx}
\usepackage{multicol}
\usepackage{caption}
\usepackage{subcaption} 
\usetikzlibrary{shapes.geometric}
\usetikzlibrary{shapes.arrows}
\usepgfplotslibrary{fillbetween}
\usepackage{array}
\usepackage{tabularx}
\usepackage{lipsum} 
 \usepackage{collcell}
\usepackage{hhline}
\usetikzlibrary{spy, calc}
\usepackage{array}
\usepackage{tabularx}
\usepackage{xspace}

\usepackage{bbding}
\usepackage{pifont}
\usepackage{diagbox}
\usepackage{booktabs}
\usepackage{makecell}
\usetikzlibrary{fit}
\usepackage{rotating}
\usepackage{filecontents}
\usepgfplotslibrary{colorbrewer}

\newcolumntype{P}[1]{>{\centering\arraybackslash}m{#1}}
\newcolumntype{R}[1]{>{\raggedright\arraybackslash}m{#1}}%

\definecolor{axisbk}{RGB}{234, 234, 242}
\pgfplotsset{colormap/Set2}

\def\addlegendimage{\csname pgfplots@addlegendimage\endcsname}

\pgfplotsset{
  /pgfplots/xlabel near ticks/.style={
     /pgfplots/every axis x label/.style={
        at={(ticklabel cs:0.5)},anchor=near ticklabel
     }
  },
  /pgfplots/ylabel near ticks/.style={
     /pgfplots/every axis y label/.style={
        at={(ticklabel cs:0.5)},rotate=90,anchor=near ticklabel}
     }
  }

\tikzset{
  fitting node/.style={
    inner sep=0pt,
    fill=none,
    draw=none,
    reset transform,
    fit={(\pgf@pathminx,\pgf@pathminy) (\pgf@pathmaxx,\pgf@pathmaxy)}
  },
  reset transform/.code={\pgftransformreset}
}

\def\layersep{1.5cm}
\def\layersepsmall{4pt}

\tikzset{cross/.style={cross out, draw, 
         minimum size=2*(#1-\pgflinewidth), 
         inner sep=0pt, outer sep=0pt}}

\makeatletter
\tikzset{
  fitting node/.style={
    inner sep=0pt,
    fill=none,
    draw=none,
    reset transform,
    fit={(\pgf@pathminx,\pgf@pathminy) (\pgf@pathmaxx,\pgf@pathmaxy)}
  },
  reset transform/.code={\pgftransformreset}
  
    }
    
\tikzset{ 
    table/.style={
        matrix of nodes,
        row sep=-\pgflinewidth,
        column sep=-\pgflinewidth,
        nodes={
            rectangle,
            draw=black,
            align=center,
            font = \small
        },
        minimum height=1.5em,
         text depth=0.5ex,
         text height=2ex,
        nodes in empty cells,
        every even row/.style={
            nodes={fill=gray!20}
        },
        row 1/.style={
            nodes={minimum width=3em}
        },
        }
}

\tikzset{
  fitting node/.style={
    inner sep=0pt,
    fill=none,
    draw=none,
    reset transform,
    fit={(\pgf@pathminx,\pgf@pathminy) (\pgf@pathmaxx,\pgf@pathmaxy)}
  },
  reset transform/.code={\pgftransformreset}
}

\pgfplotsset{
    axisStyle/.style={
    tickwidth=0mm,
    axis background/.style={fill=axisbk},
    grid=both,
    grid style={line width=.1pt, draw=white},
    legend style={fill=none, draw=none},
    legend cell align=left,
    axis line style={draw=none}
    }
}

\tikzset{
	autoencoder/.pic={
            \tikzstyle{every pin edge}=[<-,shorten <=1pt]
            \tikzstyle{neuron}=[circle,fill=black!25,minimum size=10pt,inner sep=0pt]
            \tikzstyle{input neuron}=[neuron, fill=orange!50];
            \tikzstyle{output neuron}=[neuron, fill=red!50];
            \tikzstyle{hidden neuron}=[neuron, fill=blue!50];
            \tikzstyle{annot} = [text width=4em, text centered]
        		
            \foreach \name / \y in {1,...,5}
                \node[input neuron] (I-\name) at (0,-\y) {};
            \foreach \name / \y in {1,...,3}
                \node[hidden neuron] (H-\name) at (\layersep,-\y cm-1cm) {};
            \foreach \name / \y in {1,...,5}
                \node[output neuron] (O-\name) at (2*\layersep,-\y cm) {};
        
            \foreach \source in {1,...,5}
                \foreach \dest in {1,...,3}
                    \draw[black!50] (I-\source) -- (H-\dest);
        
            \foreach \source in {1,...,3}
                \foreach \dest in {1,...,5}
                    \draw[black!50] (H-\source) -- (O-\dest);
        
            \node[annot,above = of H-1,text width=2cm] (hl) {\small hidden};
            \node[annot,above = of I-1](input) {\small input};
            \node[annot,above = of O-1](output) {\small  output};
	}
}

\tikzset{ 
    layer/.style={
    rectangle,fill=black!25,minimum width=140pt,inner sep=3pt, outer sep=0pt,minimum height=15pt,rounded corners=3pt
    }
 }
\tikzset{ 
    conv/.style={
    layer, fill=blue!30
    }
 }
 \tikzset{ 
    lrelu/.style={
    layer, fill=Yelloworange!40
    }
 }
 \tikzset{ 
    bn/.style={
    layer, fill=red!30
    }
 }
  \tikzset{ 
    dconv/.style={
    layer, fill=yellow!30
    }
 }
  \tikzset{ 
    relu/.style={
    layer, fill=Yelloworange!70
    }
 }
   \tikzset{ 
    set/.style={
    layer, fill=blue!15
    }
 }
   \tikzset{ 
    tanh/.style={
    layer, fill=pink!30
    }
 }
\tikzset{
	block1/.pic={
	\node[conv,draw](conv) {};
	\node[bn,draw,below=\layersepsmall of conv](bn) {};
	\node[lrelu,draw,below =\layersepsmall of bn](lrelu) {};	
	}
}

\tikzset{
	block2/.pic={
	\node[conv,draw](conv1) {};
	\node[bn,draw,below= \layersepsmall of conv1](bn1) {};
	\node[relu,draw,below =\layersepsmall of bn1](lrelu) {};	
	\node[conv,draw,below =\layersepsmall of lrelu](conv2) {};
	\node[bn,draw,below=\layersepsmall of conv2](bn2) {};
	}
}

\tikzset{
	block3/.pic={
	\node[dconv,draw](dconv) {};
	\node[bn,draw,below=\layersepsmall of dconv](bn) {};
	\node[relu,draw,below =\layersepsmall of bn](lrelu) {};	
	}
}

\tikzset{
	net/.pic={
	\node[layer,fill=gray,minimum width=140pt,inner sep=0,minimum height=20pt,rounded corners=2pt](l1){};
	\node[layer,fill=gray,minimum width=140pt,inner sep=0,minimum height=20pt,rounded corners=2pt,below= 3pt of l1](l2){};
	\node[layer,fill=gray,minimum width=140pt,inner sep=0,minimum height=20pt,rounded corners=2pt,below= 3pt of l2](l3){};
	}

}

\tikzset{
  annotated cuboid/.pic={
    \tikzset{%
      every edge quotes/.append style={midway, auto},
      /cuboid/.cd,
      #1
    }
    \draw [every edge/.append style={pic actions, densely dashed, opacity=.5}, pic actions]
    (0,0,0) coordinate (o) -- ++(-\cubescale*\cubex,0,0) coordinate (a) -- ++(0,-\cubescale*\cubey,0) coordinate (b) edge coordinate [pos=1] (g) ++(0,0,-\cubescale*\cubez)  -- ++(\cubescale*\cubex,0,0) coordinate (c) -- cycle
    (o) -- ++(0,0,-\cubescale*\cubez) coordinate (d) -- ++(0,-\cubescale*\cubey,0) coordinate (e) edge (g) -- (c) -- cycle
    (o) -- (a) -- ++(0,0,-\cubescale*\cubez) coordinate (f) edge (g) -- (d) -- cycle;
    ;
  },
  /cuboid/.search also={/tikz},
  /cuboid/.cd,
  width/.store in=\cubex,
  height/.store in=\cubey,
  depth/.store in=\cubez,
  units/.store in=\cubeunits,
  scale/.store in=\cubescale,
  width=10,
  height=10,
  depth=10,
  units=cm,
  scale=.1,
}

\newif\ifcuboidshade
\newif\ifcuboidemphedge

\tikzset{
  cuboid/.is family,
  cuboid,
  shiftx/.initial=0,
  shifty/.initial=0,
  dimx/.initial=3,
  dimy/.initial=3,
  dimz/.initial=3,
  scale/.initial=1,
  densityx/.initial=1,
  densityy/.initial=1,
  densityz/.initial=1,
  rotation/.initial=0,
  anglex/.initial=0,
  angley/.initial=90,
  anglez/.initial=225,
  scalex/.initial=1,
  scaley/.initial=1,
  scalez/.initial=0.5,
  front/.style={draw=black,fill=white},
  top/.style={draw=black,fill=white},
  right/.style={draw=black,fill=white},
  shade/.is if=cuboidshade,
  shadecolordark/.initial=black,
  shadecolorlight/.initial=white,
  shadeopacity/.initial=0.15,
  shadesamples/.initial=16,
  emphedge/.is if=cuboidemphedge,
  emphstyle/.style={thick},
}

\newcommand{\tikzcuboidkey}[1]{\pgfkeysvalueof{/tikz/cuboid/#1}}

\newcommand{\tikzcuboid}[1]{
    \tikzset{cuboid,#1} 
  \pgfmathsetlengthmacro{\vectorxx}{\tikzcuboidkey{scalex}*cos(\tikzcuboidkey{anglex})*28.452756}
  \pgfmathsetlengthmacro{\vectorxy}{\tikzcuboidkey{scalex}*sin(\tikzcuboidkey{anglex})*28.452756}
  \pgfmathsetlengthmacro{\vectoryx}{\tikzcuboidkey{scaley}*cos(\tikzcuboidkey{angley})*28.452756}
  \pgfmathsetlengthmacro{\vectoryy}{\tikzcuboidkey{scaley}*sin(\tikzcuboidkey{angley})*28.452756}
  \pgfmathsetlengthmacro{\vectorzx}{\tikzcuboidkey{scalez}*cos(\tikzcuboidkey{anglez})*28.452756}
  \pgfmathsetlengthmacro{\vectorzy}{\tikzcuboidkey{scalez}*sin(\tikzcuboidkey{anglez})*28.452756}
  \begin{scope}[xshift=\tikzcuboidkey{shiftx}, yshift=\tikzcuboidkey{shifty}, scale=\tikzcuboidkey{scale}, rotate=\tikzcuboidkey{rotation}, x={(\vectorxx,\vectorxy)}, y={(\vectoryx,\vectoryy)}, z={(\vectorzx,\vectorzy)}]
    \pgfmathsetmacro{\steppingx}{1/\tikzcuboidkey{densityx}}
  \pgfmathsetmacro{\steppingy}{1/\tikzcuboidkey{densityy}}
  \pgfmathsetmacro{\steppingz}{1/\tikzcuboidkey{densityz}}
  \newcommand{\dimx}{\tikzcuboidkey{dimx}}
  \newcommand{\dimy}{\tikzcuboidkey{dimy}}
  \newcommand{\dimz}{\tikzcuboidkey{dimz}}
  \pgfmathsetmacro{\secondx}{2*\steppingx}
  \pgfmathsetmacro{\secondy}{2*\steppingy}
  \pgfmathsetmacro{\secondz}{2*\steppingz}
  \ifthenelse{\equal{\dimx}{1}}
    {\foreach \x in {\steppingx,...,\dimx}}
    {\foreach \x in {\steppingx,\secondx,...,\dimx}}
  {     \ifthenelse{\equal{\dimy}{1}}
    {\foreach \y in {\steppingy,...,\dimy}}
    {\foreach \y in {\steppingy,\secondy,...,\dimy}}
    {   \pgfmathsetmacro{\lowx}{(\x-\steppingx)}
      \pgfmathsetmacro{\lowy}{(\y-\steppingy)}
      \filldraw[cuboid/front] (\lowx,\lowy,\dimz) -- (\lowx,\y,\dimz) -- (\x,\y,\dimz) -- (\x,\lowy,\dimz) -- cycle;
    }
    }
    \ifthenelse{\equal{\dimx}{1}}
    {\foreach \x in {\steppingx,...,\dimx}}
    {\foreach \x in {\steppingx,\secondx,...,\dimx}}
  { \ifthenelse{\equal{\dimz}{1}}
    {\foreach \z in {\steppingz,...,\dimz}}
    {\foreach \z in {\steppingz,\secondz,...,\dimz}}
    {   \pgfmathsetmacro{\lowx}{(\x-\steppingx)}
      \pgfmathsetmacro{\lowz}{(\z-\steppingz)}
      \filldraw[cuboid/top] (\lowx,\dimy,\lowz) -- (\lowx,\dimy,\z) -- (\x,\dimy,\z) -- (\x,\dimy,\lowz) -- cycle;
        }
    }
    \ifthenelse{\equal{\dimy}{1}}
    {\foreach \y in {\steppingy,...,\dimy}}
    {\foreach \y in {\steppingy,\secondy,...,\dimy}}
  { \ifthenelse{\equal{\dimz}{1}}
    {\foreach \z in {\steppingz,...,\dimz}}
    {\foreach \z in {\steppingz,\secondz,...,\dimz}}
    {   \pgfmathsetmacro{\lowy}{(\y-\steppingy)}
      \pgfmathsetmacro{\lowz}{(\z-\steppingz)}
      \filldraw[cuboid/right] (\dimx,\lowy,\lowz) -- (\dimx,\lowy,\z) -- (\dimx,\y,\z) -- (\dimx,\y,\lowz) -- cycle;
    }
  }
  \ifcuboidemphedge
    \draw[cuboid/emphstyle] (0,\dimy,0) -- (\dimx,\dimy,0) -- (\dimx,\dimy,\dimz) -- (0,\dimy,\dimz) -- cycle;%
    \draw[cuboid/emphstyle] (0,\dimy,\dimz) -- (0,0,\dimz) -- (\dimx,0,\dimz) -- (\dimx,\dimy,\dimz);%
    \draw[cuboid/emphstyle] (\dimx,\dimy,0) -- (\dimx,0,0) -- (\dimx,0,\dimz);%
    \fi

  \end{scope}
}

\newif\ifblackandwhitecycle
\gdef\patternnumber{0}

\pgfkeys{/tikz/.cd,
    zoombox paths/.style={
        draw=orange,
        very thick
    },
    black and white/.is choice,
    black and white/.default=static,
    black and white/static/.style={ 
        draw=white,   
        zoombox paths/.append style={
            draw=white,
            postaction={
                draw=black,
                loosely dashed
            }
        }
    },
    black and white/static/.code={
        \gdef\patternnumber{1}
    },
    black and white/cycle/.code={
        \blackandwhitecycletrue
        \gdef\patternnumber{1}
    },
    black and white pattern/.is choice,
    black and white pattern/0/.style={},
    black and white pattern/1/.style={    
            draw=white,
            postaction={
                draw=black,
                dash pattern=on 2pt off 2pt
            }
    },
    black and white pattern/2/.style={    
            draw=white,
            postaction={
                draw=black,
                dash pattern=on 4pt off 4pt
            }
    },
    black and white pattern/3/.style={    
            draw=white,
            postaction={
                draw=black,
                dash pattern=on 4pt off 4pt on 1pt off 4pt
            }
    },
    black and white pattern/4/.style={    
            draw=white,
            postaction={
                draw=black,
                dash pattern=on 4pt off 2pt on 2 pt off 2pt on 2 pt off 2pt
            }
    },
    zoomboxarray inner gap/.initial=5pt,
    zoomboxarray columns/.initial=2,
    zoomboxarray rows/.initial=2,
    subfigurename/.initial={},
    figurename/.initial={zoombox},
    zoomboxarray/.style={
        execute at begin picture={
            \begin{scope}[
                spy using outlines={%
                    zoombox paths,
                    width=0.5*\imagewidth / \pgfkeysvalueof{/tikz/zoomboxarray columns} - (\pgfkeysvalueof{/tikz/zoomboxarray columns} - 1) / \pgfkeysvalueof{/tikz/zoomboxarray columns} * \pgfkeysvalueof{/tikz/zoomboxarray inner gap} -\pgflinewidth,
                    height=\imageheight / \pgfkeysvalueof{/tikz/zoomboxarray rows} - (\pgfkeysvalueof{/tikz/zoomboxarray rows} - 1) / \pgfkeysvalueof{/tikz/zoomboxarray rows} * \pgfkeysvalueof{/tikz/zoomboxarray inner gap}-\pgflinewidth,
                    magnification=3,
                    every spy on node/.style={
                        zoombox paths
                    },
                    every spy in node/.style={
                        zoombox paths
                    }
                }
            ]
        },
        execute at end picture={
            \end{scope}
               \node at (image.north) [anchor=north,inner sep=0pt]{ \phantomimage};
               \node at (zoomboxes container.north) [anchor=north,inner sep=0pt]{ \phantomimage};
     \gdef\patternnumber{0}
        },
        spymargin/.initial=0.5em,
        zoomboxes xshift/.initial=1,
        zoomboxes right/.code=\pgfkeys{/tikz/zoomboxes xshift=1},
        zoomboxes left/.code=\pgfkeys{/tikz/zoomboxes xshift=-1},
        zoomboxes yshift/.initial=0,
        zoomboxes above/.code={
            \pgfkeys{/tikz/zoomboxes yshift=1},
            \pgfkeys{/tikz/zoomboxes xshift=0}
        },
        zoomboxes below/.code={
            \pgfkeys{/tikz/zoomboxes yshift=-1},
            \pgfkeys{/tikz/zoomboxes xshift=0}
        },
        caption margin/.initial=4ex,
    },
    adjust caption spacing/.code={},
    image container/.style={
        inner sep=0pt,
        at=(image.north),
        anchor=north,
        adjust caption spacing
    },
    zoomboxes container/.style={
        inner sep=0pt,
        at=(image.north),
        anchor=north,
        name=zoomboxes container,
        xshift=\pgfkeysvalueof{/tikz/zoomboxes xshift}*(\imagewidth+\pgfkeysvalueof{/tikz/spymargin}),
        yshift=\pgfkeysvalueof{/tikz/zoomboxes yshift}*(\imageheight+\pgfkeysvalueof{/tikz/spymargin}+\pgfkeysvalueof{/tikz/caption margin}),
        adjust caption spacing
    },
    calculate dimensions/.code={
        \pgfpointdiff{\pgfpointanchor{image}{south west} }{\pgfpointanchor{image}{north east} }
        \pgfgetlastxy{\imagewidth}{\imageheight}
        \global\let\imagewidth=\imagewidth
        \global\let\imageheight=\imageheight
        \gdef\columncount{1}
        \gdef\rowcount{1}
        
    },
    image node/.style={
        inner sep=0pt,
        name=image,
        anchor=south west,
        append after command={
            [calculate dimensions]
            node [image container,subfigurename=\pgfkeysvalueof{/tikz/figurename}-image] {\phantomimage}
            node [zoomboxes container,subfigurename=\pgfkeysvalueof{/tikz/figurename}-zoom] {\phantomimage}
        }
    },
    color code/.style={
        zoombox paths/.append style={draw=#1}
    },
    connect zoomboxes/.style={
    spy connection path={\draw[draw=none,zoombox paths] (tikzspyonnode) -- (tikzspyinnode);}
    },
    help grid code/.code={
        \begin{scope}[
                x={(image.south east)},
                y={(image.north west)},
                font=\footnotesize,
                help lines,
                overlay
            ]
            \foreach \x in {0,1,...,9} { 
                \draw(\x/10,0) -- (\x/10,1);
                \node [anchor=north] at (\x/10,0) {0.\x};
            }
            \foreach \y in {0,1,...,9} {
                \draw(0,\y/10) -- (1,\y/10);                        \node [anchor=east] at (0,\y/10) {0.\y};
            }
        \end{scope}    
    },
    help grid/.style={
        append after command={
            [help grid code]
        }
    },
}

\newcommand\phantomimage{%
    \phantom{%
        \rule{\imagewidth}{\imageheight}%
    }%
}
\newcommand\zoombox[2][]{
    \begin{scope}[zoombox paths]
        \pgfmathsetmacro\xpos{
            (\columncount-1)*(\imagewidth / \pgfkeysvalueof{/tikz/zoomboxarray columns} + \pgfkeysvalueof{/tikz/zoomboxarray inner gap} / \pgfkeysvalueof{/tikz/zoomboxarray columns} ) + \pgflinewidth
        }
        \pgfmathsetmacro\ypos{
            (\rowcount-1)*( \imageheight / \pgfkeysvalueof{/tikz/zoomboxarray rows} + \pgfkeysvalueof{/tikz/zoomboxarray inner gap} / \pgfkeysvalueof{/tikz/zoomboxarray rows} ) + 0.5*\pgflinewidth
        }
        \edef\dospy{\noexpand\spy [
            #1,
            zoombox paths/.append style={
                black and white pattern=\patternnumber
            },
            every spy on node/.append style={#1},
            x=\imagewidth,
            y=\imageheight
        ] on (#2) in node [anchor=north west] at ($(zoomboxes container.north west)+(\xpos pt,-\ypos pt)$);}
        \dospy
        \pgfmathtruncatemacro\pgfmathresult{ifthenelse(\columncount==\pgfkeysvalueof{/tikz/zoomboxarray columns},\rowcount+1,\rowcount)}
        \global\let\rowcount=\pgfmathresult
        \pgfmathtruncatemacro\pgfmathresult{ifthenelse(\columncount==\pgfkeysvalueof{/tikz/zoomboxarray columns},1,\columncount+1)}
        \global\let\columncount=\pgfmathresult
        \ifblackandwhitecycle
            \pgfmathtruncatemacro{\newpatternnumber}{\patternnumber+1}
            \global\edef\patternnumber{\newpatternnumber}
        \fi
    \end{scope}
}

\newcommand{\tsne}[1]{\begin{tikzpicture}
\begin{axis}[xtick=\empty,ytick=\empty]
\addplot[ scatter,%
        scatter/@pre marker code/.code={%
            \edef\temp{\noexpand\pgfplotspointmeta}
            \temp
            \scope[draw=none,fill=\pgfplotspointmeta]%
        },%
        scatter/@post marker code/.code={%
            \endscope
        },%
        only marks,     
        mark=*,
        point meta={TeX code symbolic={%
            \edef\pgfplotspointmeta{\thisrow{color}}%
        }},
        mark options={mark size=1.5},
    ] 
           table [x=x, y=y]{./#1};
         \legend{melanoma, nonmelanoma}
\end{axis}
\end{tikzpicture}}

\begin{document}

\title{Unsupervised  and semi-supervised learning with Categorical Generative Adversarial Networks assisted by Wasserstein distance  for dermoscopy image Classification}


%
%

\author{Xin~Yi,
Ekta~Walia,
        Paul~Babyn
\thanks{Both X. Yi and P. Babyn are with the Department
of Medical Imaging, University of Saskatchewan, Saskatoon,
SK, S7N 0W8 Canada}
\thanks{E. Walia was with University of Saskatchewan at the time of inception of this work, and at present working with Philips Canada}
}

\maketitle

\begin{abstract}
Melanoma is a curable aggressive skin cancer if detected  early. Typically, the diagnosis involves initial screening with subsequent biopsy  and   histopathological examination if necessary.  Computer aided diagnosis  offers an objective score that is independent of clinical experience and the potential to  lower the workload of a dermatologist. 
In the recent past, success of deep learning algorithms in the field of general computer vision has motivated successful application of supervised deep learning methods in computer aided melanoma recognition.  However, large quantities of labeled images are required to make further improvements on the supervised method. A good annotation generally requires  clinical and histological confirmation, which requires significant effort.   In an attempt to alleviate this constraint, we propose to use categorical generative adversarial network to automatically learn the feature representation of dermoscopy images in an unsupervised and semi-supervised manner. Thorough experiments on ISIC 2016 skin lesion challenge demonstrate that the proposed feature learning method has achieved an average precision score of 0.424 with only 140 labeled images. Moreover, the proposed method is also capable of generating real-world like dermoscopy images.

\begin{IEEEkeywords}
Dermoscopy, Categorical Generative Adversarial Networks, Unsupervised learning, Semi-supervised learning, Deep Learning , Melanoma classification 
\end{IEEEkeywords}
\end{abstract}

\section{Introduction}
Skin cancer is the most prevalent cancer in Canada, with the incidence number almost equal to the four major cancers (lung, breast, colorectal, prostate) combined. Melanoma, as one of the two major skin cancer types, is the most deadly form with a 5-year survival rate of about 14\% if detected late~\cite{esteva2017dermatologist}. According to the Canadian Cancer Statistics 2014, the incidence rates of melanoma has increased by 2.05\% per year for both sex combined between 1992 to 2013~\cite{statistics2014canadian}.  An estimate of  nearly 7300 people are expected to be diagnosed with melanoma and over 1200 are expected to die in 2017~\cite{statistics2017canadian}. This cancer arises when the pigment containing cells named melanocytes start to multiply without control and form malignant tumours.  Despite its aggressiveness, this cancer is curable  if detected early with a 5-year survival rate   over 99\%~\cite{esteva2017dermatologist}. Therefore, a convenient and accurate method for early diagnosis of melanoma is of great practical impact.

Dermoscopy is a non-invasive skin imaging technique that has been widely adopted by dermatologists for the initial screening of skin disease. Utilizing a high magnification factor, it can minimize skin reflection interference, offering a better view of the skin lesion as compared to naked eyes~\cite{silveira2009comparison}.  However, the median wait time to see a dermatologist  is over three months  in Canada~\cite{allianceskin}. It would be of great value to develop an automated melanoma recognition system based on dermoscopy images to produce an accountable screening result in a more timely manner. The significance of automated image analysis and recognition for identification of melanoma has been highlighted in recent publications~\cite{fornaciali2016towards, stoecker2013automatic, mishra2016overview,ali2012systematic}.

\begin{figure}[tbp]
\centering
\resizebox{0.5\textwidth}{!}{
\begin{tikzpicture} [
    auto,
    line/.style     = { draw, thick, ->, shorten >=2pt,shorten <=2pt },
    every node/.append style={font=\Large}
  ]
 \matrix [column sep=2mm, row sep=2mm,ampersand replacement=\&] {
		 \node (p11)[inner sep=0] at (0,0){\includegraphics[width=0.33\textwidth, height=0.22\textwidth]{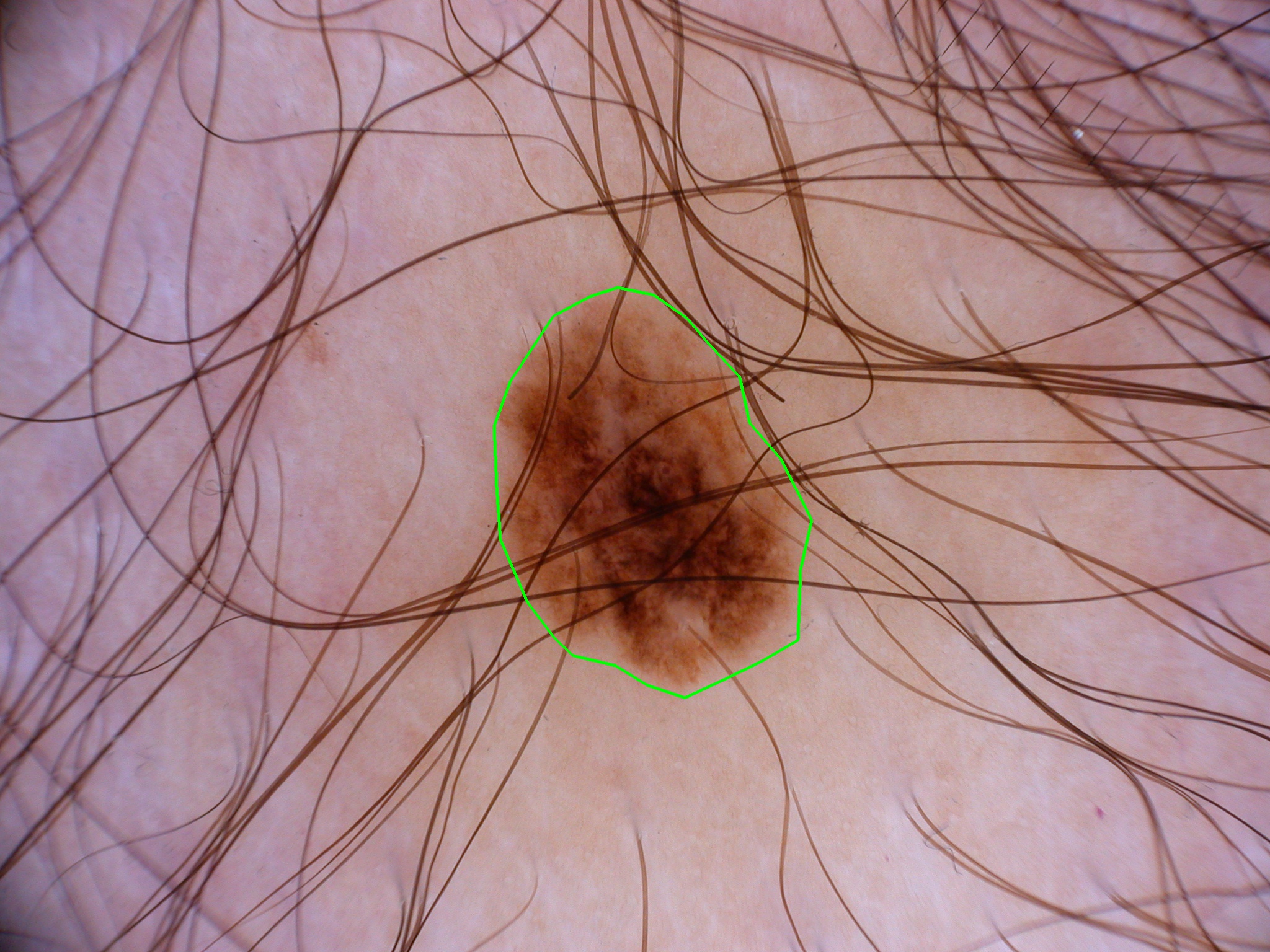}};     \&
		 \node (p12)[inner sep=0] at (0,0){\includegraphics[width=0.33\textwidth, height=0.22\textwidth]{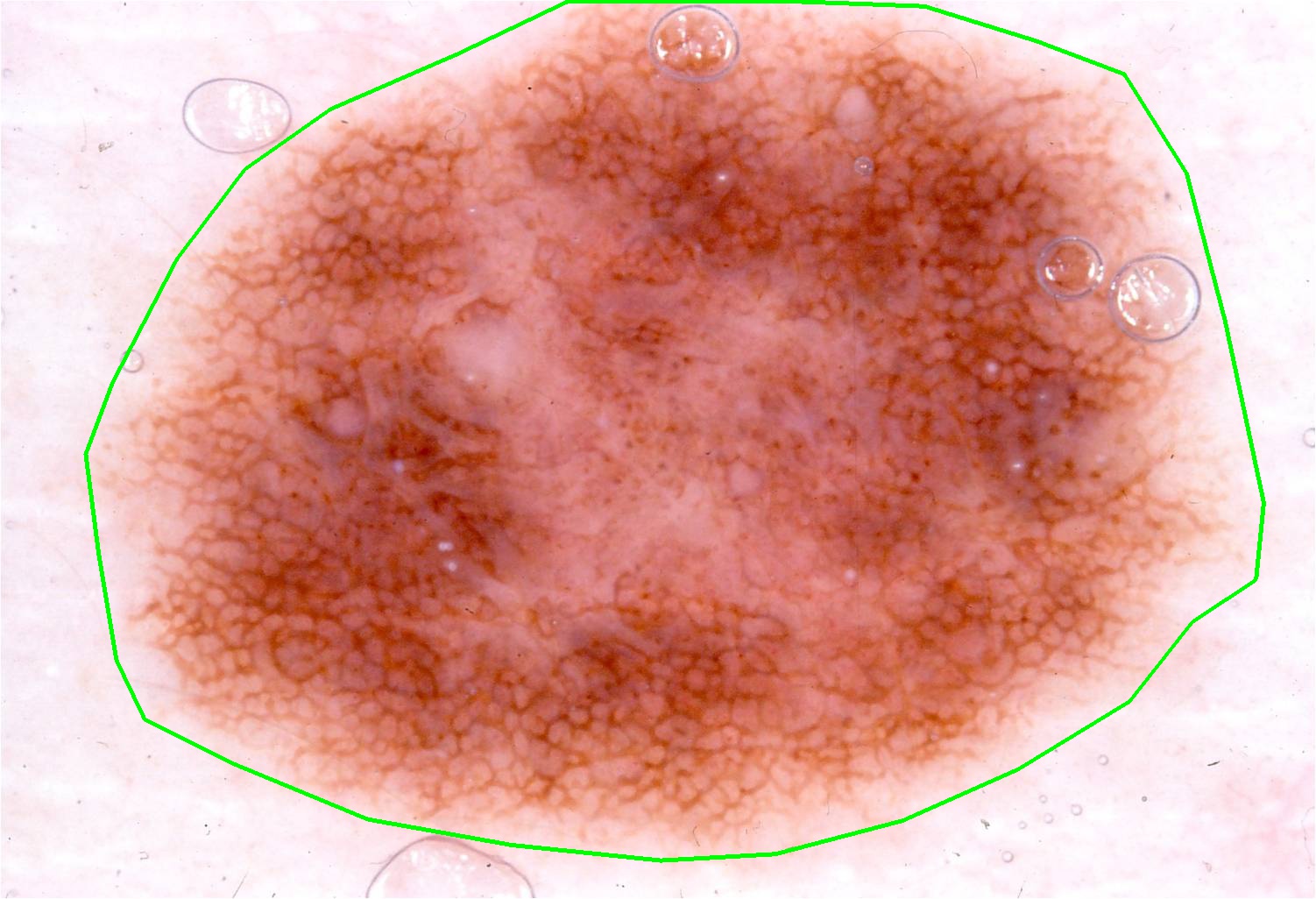}};     \&
		 \node (p13)[inner sep=0] at (0,0){\includegraphics[width=0.33\textwidth, height=0.22\textwidth]{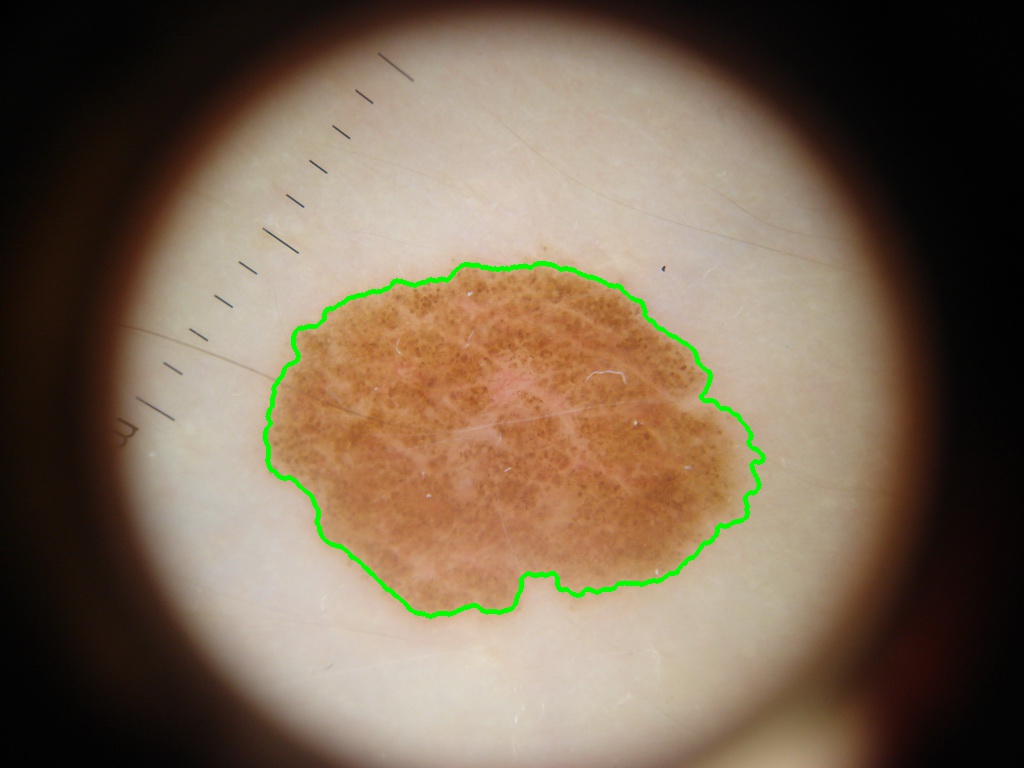}};     \&\\
                       };
  \begin{scope} [every path/.style=line]
     \node[anchor=north] at (p11.south) {Hair};
     \node[anchor=north] at (p12.south) {Air bubble};
      \node[anchor=north] at (p13.south) {Ruler};
  \end{scope}
\end{tikzpicture}
}
\caption{Exemplar dermoscopy images with artifacts such as hair, air bubble and ruler. 
}
\label{artifact}
\end{figure}
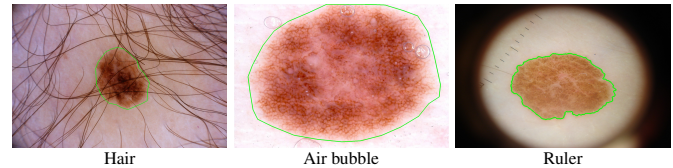

\begin{figure}[tbp]
\centering
\resizebox{0.5\textwidth}{!}{
\begin{tikzpicture} [
    auto,
    line/.style     = { draw, thick, ->, shorten >=2pt,shorten <=2pt },
    every node/.append style={font=\large}
  ]
 \matrix [column sep=0.3mm, row sep=2mm,ampersand replacement=\&] {
		 \node (p11)[inner sep=0] at (0,0){\includegraphics[width=0.8\textwidth]{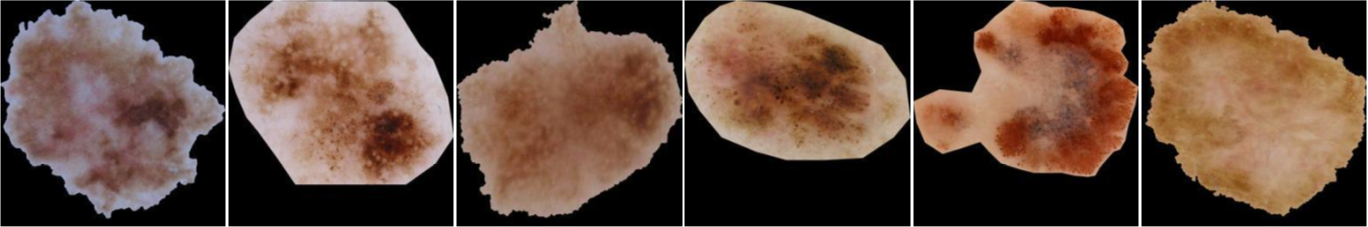}};     \&\\
		 \node (p21)[inner sep=0] at (0,0){\includegraphics[width=0.8\textwidth]{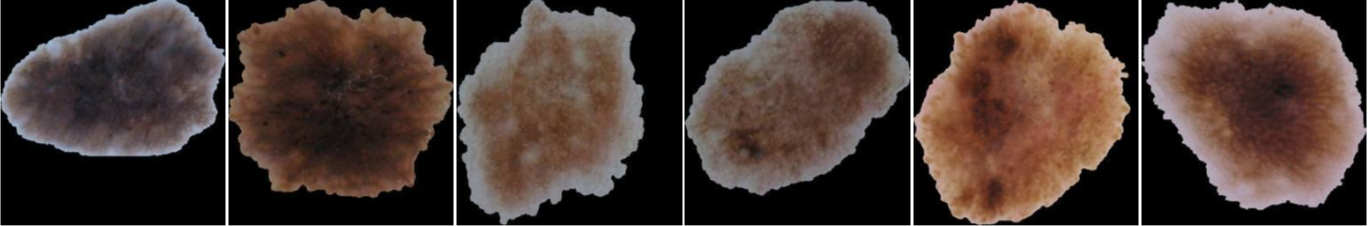}};     \&\\
                       };
  \begin{scope} [every path/.style=line]
     \node[anchor=east,rotate=90,yshift=3mm,xshift=7mm] at (p11.west) {\small Melanoma};
     \node[anchor=east,rotate=90,yshift=3mm,xshift=5mm] at (p21.west) {\small Benign};
  \end{scope}
\end{tikzpicture}
}
\caption{Sample melanoma and benign images. Background is segmented to only show the skin lesion.}
\label{sampleimgs}
\end{figure}
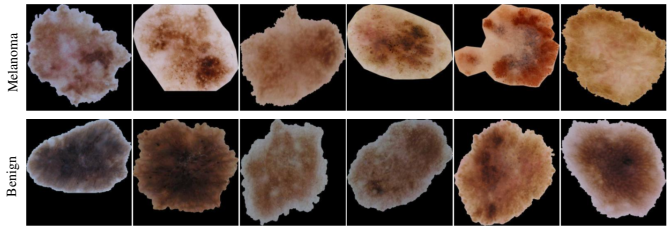

There are a couple of challenges  in the development of a dermoscopy image based automation system. First, there is no discernible boundary between the normal and lesion skin which makes the segmentation of skin lesions difficult, let alone the artifacts that can be  seen in the image, e.g. hair, bubble and ruler. Sample images with these three artifacts can be seen in Figure~\ref{artifact}. Another big problem is the large intra-class variation (among melanomas) and small inter-class variation (melanoma v.s. benign) as demonstrated in Figure~\ref{sampleimgs}. The skin lesions are segmented from the surrounding region to give a better illustration.

Recently, supervised learning with Convolutional Neural Networks (CNN) has provided excellent results for classification of skin cancer~\cite{esteva2017dermatologist}. Dermotologist-level performance was claimed with 129,450 clinical images, including 3,374 labeled dermoscopy images used in the training. However, the authors stated that their method is constrained by data and the application can be extended to more visual conditions if sufficient labeled training examples exist. Similar claim has been made in~\cite{ali2012systematic}   where Ali et al.  stated that the method they presented is constrained by data and needs sufficient training examples to succeed in the classification of various skin conditions. 

Progressive performance seems to be straightforward  by simply acquiring more labeled images. However, obtaining large volumes of labeled training data  is time consuming and requires extensive  expertise. In natural image field, we could rely on Amazon Turk to outsource the labeling task but this only suffices for general object categories. Fine-grained image labeling such as differentiating bird species would also require domain specific knowledge, not to mention dermoscopy images that require years of medical training and sometimes demand histopathology consultation.  Moreover, privacy issues would further limit the number of qualified candidates.

Nonetheless, there is substantial amount of unlabeled images that have not been utilized, and their  acquisition is relatively easier and inexpensive. Therefore, in this work, we explore unsupervised and semi-supervised  learning techniques for dermoscopy image classification, in particular melanoma classification. Our proposed method   combined categorical generative adversarial network (catGAN)~\cite{springenberg2015unsupervised} and Wasserstein distance (also known as earth mover's distance in computer science)~\cite{gulrajani2017improved} for  its automatic feature learning. Evaluation on the ISIC skin lesion challenge 2016 dataset has shown promising results. We refer the proposed method as catWGAN and its detailed description can be found in Section~\ref{catwgan}. Note that in this work, we mainly assess the feature learning capability of the network. As shape statistics of skin lesion plays an important role in the decision making of dermatologist,  the skin lesions were segmented with ground truth segmentation maps to avoid interference of backgrounds and possible confusion that could be introduced by various segmentation algorithms. 

 \section{Related works}
High dimensional image data is usually assumed to lie on a lower dimensional manifold where the original data can be projected  to get a more compact feature representation~\cite{chapelle2009semi}. In automatic classification systems, this change of data representation can not only save  computation, but also makes the system robust to  potential transformations encountered in the real-world, such as illumination, rotation, scale change, and translation. Based on the strategies of how the features are designed, we broadly categorize the existing literature into two groups,  one centred on use of hand-crafted features and the other focused on automatically learnt features.
 
 \subsection{Hand-crafted features}
In the pre-deep learning era, hand-crafted features  based on shape~\cite{friedman1985early, celebi2009lesion}, color~\cite{stanley2007relative, cheng2008skin, barata2015improving}  and texture~\cite{ballerini2013color, majtner2016efficient} of skin lesion  played a key role in automatic melanoma classification systems.  
The features  used are very broad and are generally adopted from the  traditional computer vision literature. For example,   Garnavi et al. used a wavelet decomposition based texture feature extraction method  in conjunction with other geometric and border based features of the lesions~\cite{garnavi2012computer}. Jafari et al.  combined asymmetry, colour and border assessment features with a Support Vector Machine (SVM) classifier to automate melanoma detection~\cite{jafari2016automatic}.  Integration of local and global features is also exploited in some computer-aided diagnosis systems~\cite{barata2015melanoma, barata2014two}.  Among various ways of fusing features from different sources, bag of visual word (BoW) is probably the most popular method to aggregate the low-level features  into so-called mid-level features which are more robust to image variations. This has been adopted in many dermoscopy image classification works~\cite{barata2014bag, alfed2016improving,situ2008malignant}. A more in-depth review of these traditional techniques can be found in~\cite{pathan2018techniques}.

\subsection{Automatically learned features with deep neural networks}
Recently, deep learning based algorithms have quickly dominated most vision based tasks. These advances have been quickly brought into  the field of computer aided diagnosis, including retinopathy~\cite{gulshan2016development}, breast cancer diagnosis~\cite{cruz2017accurate}, pulmonary nodules detection~\cite{ciompi2017towards} and the focus of our work, melanoma image classification.  

\subsubsection{Supervised learning:} 

Previous deep learning work in melanoma classification has prevailingly used fully supervised learning for  feature extraction and then attach a classifier such as random forest on top for classification. Constrained by the fact that not enough labeled samples are available to fully train the neural network, a large portion of related works use a transfer learning scheme by fine-tuning a pre-trained neural network. The underlying assumption is that the cascade level of features learned with natural images could also be beneficial for medical images especially those features learnt in  the first few layers of the network which are mainly edges and some other simple image structures. The training could then focus on the deeper layers leaving the shallow layers untouched.

Codella et al. extracted features  from a pre-trained CNN and further combined traditional sparse coding features  for melanoma recognition~\cite{codella2015deep}. Liao fine-tuned three different pre-trained CNNs (VGG15, VGG19, GoogleNet)  for universal skin disease classification~\cite{liao2016deep}. The effectiveness of transfer learning was also manifested in other similar works that adopts the same fine-tuning strategy~\cite{lopez2017skin, yu2017automated, gutman2016skin}.

%
\subsubsection{Unsupervised and semi-supervised learning:}

 Unsupervised and semi-superviesd learning method is not  commonly used in this field mostly due to its modest performance. The only work we have found, uses a stacked sparse autoencoder  to learn hierarchical level of features for classification of skin lesion images~\cite{sabbaghi2016deep}.  However, we believe unsupervised and semi-supervised methods will play an important role in solving medical imaging problems due to the scarcity of labeled medical datasets. Here we briefly review contemporary deep learning based semi-supervised learning methods that are not only  scalable to large quantities of unlabeled images but are also capable of simultaneously performing the unsupervised and supervised learning task (opposed to unsupervised pre-training followed by supervised fine-tuning). For a more in-depth review of traditional semi-supervised learning methods, we refer the reader to these two works~\cite{schoneveld2017semi, chapelle2009semi}.
 
Recent methods typically involve training of  a feed-forward classifier together with some auxiliary unsupervised tasks, hoping the learnt features would generalize better. The most common unsupervised tasks includes  minimizing reconstruction error of inputs~\cite{hinton2006reducing, kingma2013auto, maaloe2016auxiliary} or learnt intermediate representations~\cite{rasmus2015semi}, encourage model invariance of input data perturbations~\cite{dosovitskiy2014discriminative, sajjadi2016regularization, miyato2017virtual},  or some ways of data embedding~\cite{ranzato2008semi, weston2012deep}.

%
%

Generative modelling, which is a branch of unsupervised learning, has seen rapid progress during the last several years. Generative adversarial network (GAN)~\cite{goodfellow2014generative}, in particular, has received  attention due to its capability of generating synthetic real-world like samples. The extension of GAN into semi-supervised learning  has achieved state-of-the-art results on CIFAR10~\cite{donahue2016adversarial, miyato2017virtual}, MNIST~\cite{springenberg2015unsupervised} and SVHN~\cite{dai2017good, li2017triple}.   In this work, we explored one type of generative model, named categorial GAN (catGAN) for unsupervised and semi-supervised learning. catGAN is a generalization of the  GAN model to multiple classes where the adversarial loss used involves information maximization.    Since the original catGAN training is not stable, we further adopted the Wasserstein distance for assistance and we refer the proposed approach as  catWGAN.  We have shown that with only 140 labeled samples,  the learned feature outperforms simple hand-crafted features and baseline denoising autoencoder by a large margin. The model can also synthesize real-world like dermoscopy images (Figure~\ref{large}) that could potentially be further used as a source of data augmentation and potentially for training of dermatologists. As far as we know, this is the first work that applied GAN based semi-supervised learning on melanoma classification.


\section{Methodology}\label{catwgan}
\begin{figure*}[htp]
\resizebox{0.9\textwidth}{!}{
\begin{tikzpicture}
\node (input)[inner sep=0,label={[align=center]below: Noise\\ $z\sim p(z)$}] at (0,0){
    \tikz{\node (im)[anchor=south west,inner sep=0,opacity=1] at (0,0){\includegraphics[width=1.6cm, height=1.6cm]{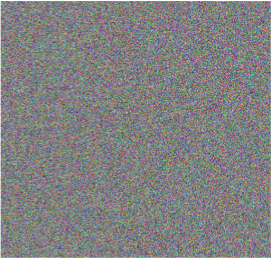}};
	}
 };  

\node[scale=0.8,rectangle,fill=gray!20,label=left:G,node distance=10pt,rotate=90, right= 0.5cm of input, anchor=north](G) { \tikz[every node/.append style={scale=0.3}]{\pic[solid](net1) {net};}};

\node (output)[inner sep=0,label={[align=center]below: Generated images\\ $\hat{x} \sim p(\hat{x})$},right= of G.center ]{\tikz{
\begin{scope}
    \node (im)[anchor=south west,inner sep=0,opacity=0.3] at (0,0){\includegraphics[width=1.6cm]{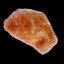}};
         \draw[thick, dashed,step=0.2cm,draw=black,fill=yellow!50,very thin,opacity=0.5] (0,0) rectangle (1.6,1.6);
\end{scope}
   \begin{scope}[xshift=0.3cm,yshift=0.3cm]
    \node (im)[anchor=south west,inner sep=0,opacity=0.3] at (0,0){\includegraphics[width=1.6cm]{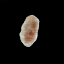}};
              \draw[ thick, dashed, step=0.2cm,draw=black,fill=blue!50,opacity=0.5] (0,0) rectangle (1.6,1.6) ;
    \end{scope}
         
	}
};

\node (train)[inner sep=3pt,below=55pt of output,label={[align=center]below:Training sample} ]{\tikz{
\begin{scope}
    \node (im)[anchor=south west,inner sep=0,opacity=0.3,label={[align=center]below: Real images\\ $x \sim p_{data}(x)$}] at (0,0){\includegraphics[width=1.6cm]{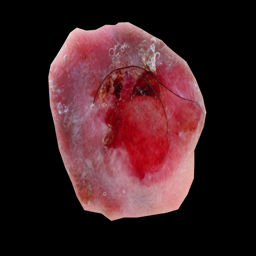}};
         \draw[step=0.2cm,draw=black,fill=yellow!50,very thin,opacity=0.5] (0,0) rectangle (1.6,1.6);
\end{scope}
   \begin{scope}[xshift=0.3cm,yshift=0.3cm]
    \node (im)[anchor=south west,inner sep=0,opacity=0.3] at (0,0){\includegraphics[width=1.6cm]{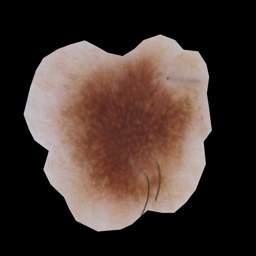}};
              \draw[step=0.2cm,draw=black,fill=blue!50,opacity=0.5] (0,0) rectangle (1.6,1.6) ;
    \end{scope}
         
	}
};  
\draw[draw=gray!70]($(train.south west)+(-5pt, 0pt)$)  rectangle ($(output.north east)+(10pt, 5pt)$);

\node(inputset1)[rectangle,fill=gray!10,minimum width=50pt,  right=45pt of output]{\tikz{
    	\node(m)[label={[align=center]below: Melanoma}]{\tikz{ 
		\draw[ultra thick, dashed,step=0.2cm,draw=black,fill=yellow!50,very thin,opacity=0.5] (0,0) rectangle (1.6,1.6);
		\draw[step=0.2cm,draw=black,fill=yellow!50,very thin,opacity=0.5] (0.3,0.3) rectangle (1.9,1.9) ;
		}};
	\node(nm)[ right= 0.5cm of m, label={[align=center]below: Benign}]{\tikz{
		\draw[ultra thick, dashed, step=0.2cm,draw=black,fill=blue!50,very thin,opacity=0.5] (0,0) rectangle (1.6,1.6);
		\draw[ step=0.2cm,draw=black,fill=blue!50,very thin,opacity=0.5] (0.3,0.3) rectangle (1.9,1.9) ;
	}};
	}};

\node[scale=0.8,rectangle,fill=gray!20,label=left:  $\text{D}_1$,node distance=10pt,rotate=90, right= 0.5cm of inputset1, anchor=north](S2) { \tikz[every node/.append style={scale=0.3}]{\pic[solid](net2) {net};}};
\draw[->,>=stealth](inputset1) -- (S2)--++(30pt,0) node[text width=1.2cm,anchor=west]{
\begin{tikzpicture}
\begin{scope}[yshift=1cm]
\draw[fill=blue!50] (0,0) rectangle (0.2cm, 1cm);
\draw[fill=yellow!50] (0.4cm,0) rectangle (0.6cm, 0.2cm) node[yshift=0.2cm]{or};
\draw[fill=blue!50] (0.9cm,0) rectangle (1.1cm, 0.2cm);
\draw[fill=yellow!50] (1.3cm,0) rectangle (1.5cm, 1cm);
\node(t1)[text width=3cm] at (-0.1cm,-0.3cm){for real sample};
\end{scope}
\begin{scope}[yshift=-0.5cm,xshift=0.5cm]
\draw[fill=blue!50] (0,0) rectangle (0.2cm, 0.5cm);
\draw[fill=yellow!50] (0.4cm,0) rectangle (0.6cm, 0.5cm);
\node[text width=3cm, below=1cm of t1] {for generated sample};
\end{scope}
\end{tikzpicture}

};

\node(inputset2)[rectangle,fill=gray!10,minimum width=50pt,below=45pt of inputset1]{\tikz{
    	\node(m)[label={[align=center]below: Real \\ $x$}]{\tikz{ 
		\draw[step=0.2cm,draw=black,fill=yellow!50,very thin,opacity=0.5] (0,0) rectangle (1.6,1.6);
		\draw[step=0.2cm,draw=black,fill=blue!50,very thin,opacity=0.5] (0.3,0.3) rectangle (1.9,1.9) ;
		}};
	\node(nm)[ right= 0.5cm of m, label={[align=center]below: Generated\\$\hat{x}$}]{\tikz{
		\draw[thick, dashed, step=0.2cm,draw=black,fill=yellow!50,very thin,opacity=0.5] (0,0) rectangle (1.6,1.6);
		\draw[thick, dashed, step=0.2cm,draw=black,fill=blue!50,very thin,opacity=0.5] (0.3,0.3) rectangle (1.9,1.9) ;
	}};
	}};

\node[scale=0.8,rectangle,fill=gray!20,label=left: $\text{D}_2$,node distance=10pt,rotate=90, right= 0.5cm of inputset2, anchor=north](D2) { \tikz[every node/.append style={scale=0.3}]{\pic[solid](net2) {net};}};
\draw[->,>=stealth](inputset2) -- (D2)--++(30pt,0) node[text width=1cm,anchor=west]{ Real or Fake? \\$\mathcal{L}_{adv}(G,D)$};

\begin{scope}[on background layer]
    \draw[](input) -- (output);
    \draw[](output.east) -++(10pt,0) node[inner sep=0](corner){} |- (inputset1.west);
    \draw[](output.east) -- (corner) |- (inputset2.west);
\end{scope}

\draw[orange!50, line width=5pt,->,>=stealth] ($(input)+(0,-2cm)$) -- ++ (13cm,0) node[black,anchor=west]{catGAN};
\draw[red!50, line width=5pt,->,>=stealth] ($(input)+(0,-2.3cm)$)-- ($(output.east)+(0,-2.3cm)$)  -- ++ (18pt,0) |- ($(inputset2.west)+(0,-2.2cm)$) -- ++(7.1cm,0) node[black,anchor=west]{WGAN};

\node (caption)[ inner sep=0pt,below=of input ,xshift=15pt,yshift=-55pt]{\tikz{
    	\draw[step=0.2cm,draw=black, thick] (0,0) rectangle (0.4,0.4) node[anchor=west,yshift=-5pt,xshift=3pt]{Real};
    	\draw[step=0.2cm,draw=black, thick,yshift=-0.5cm,dashed] (0,0) rectangle (0.4,0.4)node[anchor=west,yshift=-5pt,xshift=3pt]{Generated};
	\draw[step=0.2cm,fill=yellow!50,yshift=-1cm,draw=none] (0,0) rectangle (0.4,0.4)node[anchor=west,yshift=-5pt,xshift=3pt]{Melanoma};
    	\draw[step=0.2cm,fill=blue!50,yshift=-1.5cm,draw=none] (0,0) rectangle (0.4,0.4)node[anchor=west,yshift=-5pt,xshift=3pt]{Benign};

}
};

\end{tikzpicture}
}

\caption{Overview of catWGAN. G is the generator that is responsible for synthetic sample generation. The output $\text{D}_1$ are confidence values for the two classes we are interested in, melanoma and benign. The orange line is used to depict the catGAN architecture and the red line depicts the WGAN architecture used to assist the training of catGAN.}
\label{overview}
\end{figure*}
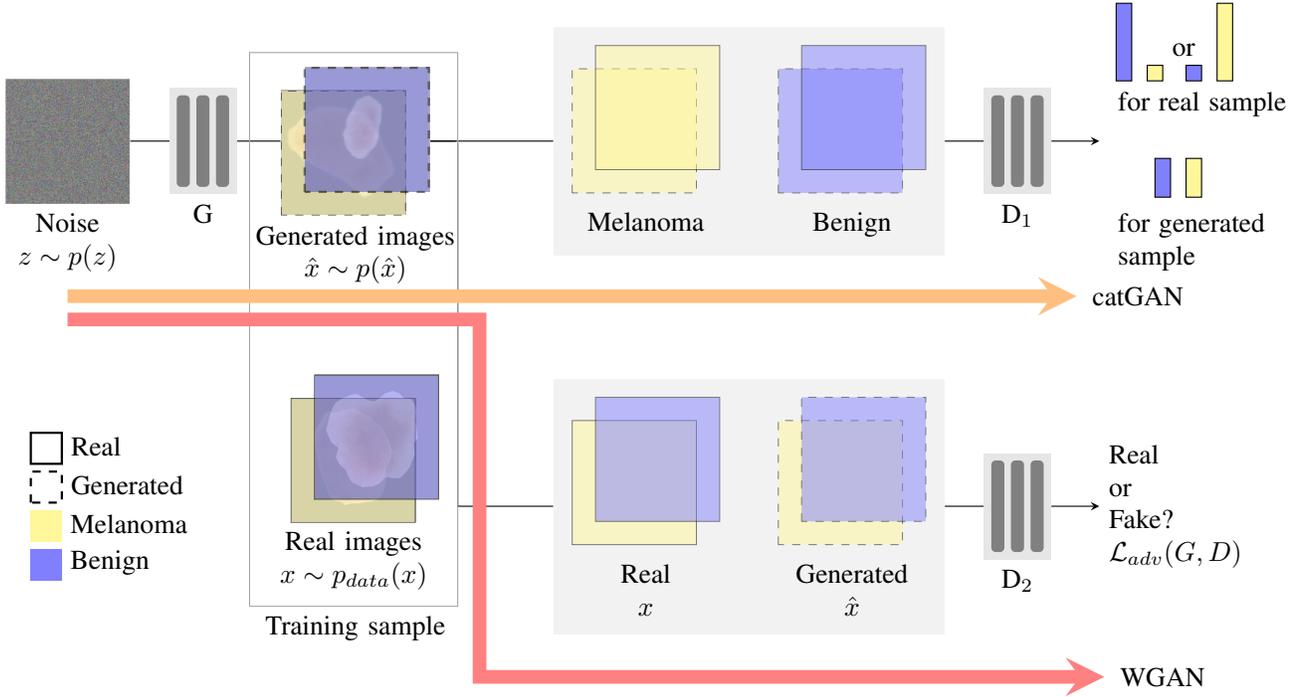

The traditional GAN is a generative model that implicitly estimates the sample distribution so that we can directly sample from the model. It consists of two types of networks: the generator G that generates synthetic samples from pure noise and the discriminator $\text{D}_1$ that   differentiates between real and generated samples. G and $\text{D}_1$ update themselves alternatively during the training process with contradictory objective. catGAN adopts the general framework of GAN but modifies the objective of $\text{D}_1$ in a way that rather than classifying the input sample as ``real'' or ``fake'', it  outputs confidence values of input belonging to each one of the underlying classes.  Figure~\ref{overview} following the orange line illustrates this process.

Let  $G: \mathbb{R}^d \rightarrow \mathbb{R}^{m\times n}$ denote a mapping from random noise $z \sim p(z)$ to a generated image sample $\hat{x}$,  and  $D_1: \mathbb{R}^{m\times n} \rightarrow \mathbb{R}^2$ denote a mapping from an input sample image  to its predicted label distribution $y$. The input could be either real sample $x \sim p_r(x)$ or generated fake sample $\hat{x} \sim p_g(\hat{x})$.
In unsupervised setting, the performance of $\text{D}_1$ was measured by the peakedness of the output label distribution using entropy.  The overall objective of $\text{D}_1$ and $\text{G}$ in catGAN formulation can be expressed mathematically as:
\begin{equation}
\begin{split}
\mathcal{L}^{\text{catGAN}}_{D_1} = \max_{D_1} &\text{ }H_{x\sim p_r{(x)}}[p(y\mid D_1)]- \underbrace{\mathbb{E}_{x\sim p_r{(x)}}\big[H[p(y\mid x,D_1)]\big]}_{S_r}\\&+ \underbrace{\mathbb{E}_{z \sim p(z)}\big[H[p(y\mid G(z),D_1)]\big]}_{S_g}, 
\end{split}
\label{e1}
\end{equation}

\begin{equation}
\mathcal{L}^{\text{catGAN}}_G = \min_{G} -H_G[p(y\mid D_1)]+ \underbrace{\mathbb{E}_{z\sim p(z)}\big[H[p(y\mid G(z),D_1)]\big]}_{S_g}, 
\end{equation}
where $H_{x\sim p_r{(x)}}[p(y\mid D_1)]$ and $H_G[p(y\mid D_1)]$ is the entropy of the marginalized class distribution  over  real and generated samples respectively. These two entropies were maximized to ensure equal usage of samples from both classes. The second term in $\mathcal{L}^{\text{catGAN}}_{D_1} $ is the estimated entropy of the predicted class distribution over real samples, that $\text{D}_1$ tries to minimize. The $S_g$ term as appearing in both $\mathcal{L}^{\text{catGAN}}_{D_1}$ and $\mathcal{L}^{\text{catGAN}}_G$ is the estimated entropy of the predicted class distribution over generated samples, over which $\text{D}_1$ tries to maximize and G tries to minimize. $S_r$ and $S_g$ are used to denote these two terms for future reference.

Similar to traditional GANs, stabilizing the training of catGAN so that neither G or $\text{D}_1$ is  overpowered by the other is a significant issue, as already pointed out in the original catGAN paper. The generator would stop improving when the discriminator becomes too strong, where the loss of the discriminator  gets saturated  and leads to zero gradients for updating G.  Although by adopting the DCGAN architecture~\cite{radford2015unsupervised}, the likelihood of model collapse decreased a lot,  we did observe this unstable phenomenon from time to time with different initializations. Therefore,  to cope with this problem,  we further employed a second discriminator $\text{D}_2$ with Wasserstein distance~\cite{arjovsky2017wasserstein} for assistance as shown in Figure~\ref{overview} following the red line. 

Springenberg has compared the catGAN formulation to the regularized information maximization (RIM) framework, and found that the generator of catGAN can be thought of as an adaptively learned regularizer for its discriminator~\cite{springenberg2015unsupervised}. Under this perspective, the better the generated sample becomes, the more robust the $\text{D}_1$ becomes to the adversarial samples. This  constitutes another part of the motivation of the integration of the second discriminator. When $\text{D}_1$ failed to supply enough gradients to update G, $\text{D}_2$ will still offer gradients to help G catch up.

The objective of $\text{D}_2$ is to differentiate between real and generated fake samples as in the traditional GAN without worrying about the composition of underlying classes. Traditional GANs minimize a $f$-divergence between the real data distribution and the generated data distribution~\cite{goodfellow2014generative, nowozin2016f}. Since $f$-divergence is a function of the density ratio, it would either become zero or infinite when the support of the two distributions do not overlap.  Using Wasserstein distance mitigates this problem by assuming a Lipschitz constraint on the discriminator\footnote{In some publications, this is called critic because the output is no longer a confidence value of real or generated sample but a real number. We stick to the name of discriminator for the sake of consistency in this paper.}.  Gradient penalty  was employed here for the training of $\text{D}_2$, as it was shown to be beneficial for the training of WGAN with various architectures~\cite{gulrajani2017improved}. 

The loss can be expressed as follows in the WGAN formulation
\begin{equation}
\begin{split}
\mathcal{L}^{\text{WGAN}}_{D_2} =\max_{D_2} & \text{ }-\mathbb{E}_{x\sim p_r(x)}[(D_2(x)] + \mathbb{E}_{\hat{x}\sim p_g(\hat{x})}[D_2(\hat{x})] \\
					&-\lambda\mathbb{E}_{\hat{x}\sim p_g{(\hat{x})}}[(||\nabla_{\hat{x}} D_2(\hat{x}) ||)^2], 
\end{split}
\label{critic_loss}
\end{equation}
\begin{equation}
\mathcal{L}^{\text{WGAN}}_G =\min_G \text{ }  \mathbb{E}_{z\sim p(z)}[D_2(G(z))] ,
\end{equation}
where $\lambda$ is the weight of the gradient penalty. Combining equation (1) to (4) into the same framework, we have the full unsupervised objective in our case as:
\begin{equation}
\begin{split}
\mathcal{L}_{\text{unsuper}}(G, D_1,D_2) =  \min_G \max_{D_1,D_2}  \text{ } \mathcal{L}_G^{\text{catGAN}} +\alpha \mathcal{L}_G^{\text{WGAN}}\\+ \mathcal{L}^{\text{catGAN}}_{D_1} + \mathcal{L}^{\text{WGAN}}_{D_2} ,
\end{split}
\end{equation}	
where $\alpha$ weights the influence of $\text{D}_1$ and $\text{D}_2$ to G. The negative of the first two terms of equation~\ref{critic_loss} is the Wasserstein distance between the generated distribution and the real sample distribution.

The extension to semi-supervised training is straightforward by incorporating the cross entropy (CE) loss  for the labeled samples in equation~\ref{e1}, so that the loss for $\text{D}_1$ becomes:
\begin{equation}
\begin{split}
\mathcal{L}^{\text{catGAN}}_{D_1} = \max_{D_1} \text{ }&H_{x\sim p_{(x)}}[p(y\mid D_1)]- \underbrace{\mathbb{E}_{x\sim p_r{(x)}}\big[H[p(y\mid x,D_1)]\big]}_{S_r}\\&+ \underbrace{\mathbb{E}_{z\sim p(z)}\big[H[p(y\mid G(z),D_1)]\big]}_{S_g} \\&+ \lambda \mathbb{E}_{(x,\hat{y})\sim \mathcal{X}^l}[CE\big[\hat{y}, p(y|x,D_1)]\big],
\end{split}
\label{e2}
\end{equation}
where $\mathcal{X}^l$ is the set of labeled samples $\{(x^1, \hat{y}^1),(x^2, \hat{y}^2)\} \cdots(x^l, \hat{y}^l)\}$.
\begin{figure}[!t]
\centering
\begin{tikzpicture} [
    auto,
    line/.style     = { draw, thick, ->, shorten >=2pt,shorten <=2pt },
    every node/.append style={font=\large}
  ]
 \matrix [column sep=2mm, row sep=6mm,ampersand replacement=\&] {
		 \node (p11)[inner sep=0] at (0,0){\includegraphics[width=0.49\textwidth]{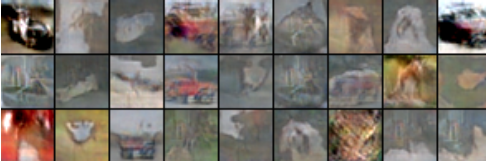}};     \&\\
		 \node (p12)[inner sep=0] at (0,0){\includegraphics[width=0.49\textwidth]{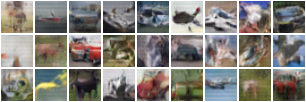}};     \&\\
                       };
  \begin{scope} [every path/.style=line]
     \node[anchor=north] at (p11.south) {\small (a)};
     \node[anchor=north] at (p12.south)  {\small (b)};

  \end{scope}
\end{tikzpicture}

\caption{Generated images on CIFAR 10. (a) is the generated result from the original catGAN architecture (directly cropped from the original paper). (b) shows the generated results from the proposed catGAN architecture (no $\text{D}_2$ in this experiment). }
\label{compare}
\end{figure}
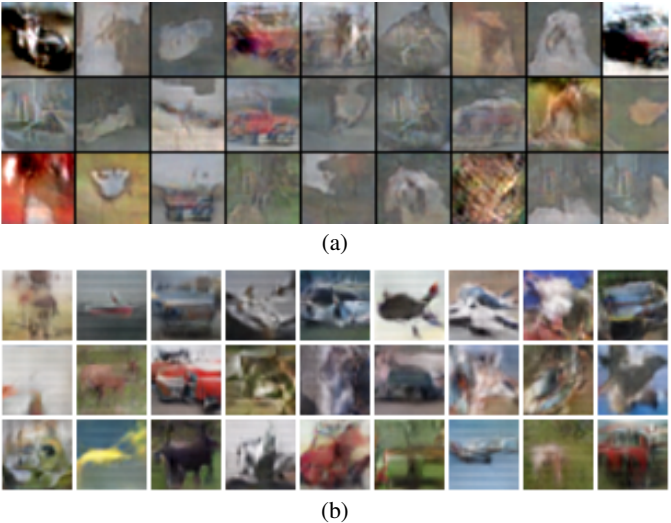

\subsection{Network architectures}
The  architectures described below were used to generate images of size $64 \times 64$. For the catGAN part (G and $\text{D}_1$), we found the original architecture tends to produce low contrast  images. Therefore, instead of using the original architecture from the catGAN paper, we customized with some key modifications. The detailed architecture can be found in Table~\ref{tb:g}. Compared with catGAN's original architecture, the proposed architecture substituted all pooling layers with stride convolution. Compared with DCGAN's architecture, we employed leaky Relu instead of Relu for the generator to avoid dead gradients and batch normalization layer was inserted after every convolution layer. Also, the feature dimension of the discriminator was further compressed from $512 \times 4 \times 4$ to $512\times1\times1$ and a 2-way softmax layer was built on top to produce the confidence values of the two classes. Figure~\ref{compare} shows the difference between the image generated from the  original and proposed catGAN architecture on CIFAR10 by just using the catGAN formulation without Wasserstein distance.

For $\text{D}_2$, we chose to restrict it to model high frequency structures as inspired by~\cite{isola2016image}. We found that by using $\text{D}_1$ alone, G could produce general structures but sometimes struggled in producing high frequency details, especially in generating high spatial resolution images. Therefore, it is reasonable to let $\text{D}_2$  focus on local image patches. The architecture of  $\text{D}_2$ can be seen in Table~\ref{tb:g}. It is a three layer fully convolutional network. The output of $\text{D}_2$ is an average over all responses.
An advantage of this architecture is that it is suitable for arbitrary input sizes and has fewer parameters. The patch size in our case is $22\times22$.

\subsection{Baseline methods}
We chose another unsupervised learning algorithm named denoising autoencoder (DAE) as the baseline for comparison. A  DAE  is one type of  neural network that learns to reconstruct a noise corrupted input. Similar to catGAN, DAE also consists of two networks, an encoder and a decoder. The encoder's job is to transform the input to a more compact feature representation that has smaller dimension than the input and the decoder  reconstructs the input from this compressed representation. The resultant feature representation should preserve all necessary information  needed for reconstruction. 

The encoder adopts $\text{D}_1$'s  architecture with the last layer chopped off and the decoder reverses the encoder's operation accordingly to output a reconstructed image of size $64\times 64$. In this manner, the learnt feature representation would have the same dimension as that of the proposed catWGAN for a fair comparison.

 To demonstrate the effectiveness of the proposed method, we also compared   two simple hand-crafted features that are commonly used in melanoma classification: edge histogram and color histogram~\cite{abedini2015generalized, codella2016deep}

\begin{table}[htp]
\tiny
\centering
\small
\begin{tabular}{cc}
		\multicolumn{2}{c}{Generator (G)}\\\toprule
		Layer				&	Activation Size 	\\\midrule
		Input noise				&	$z\in \mathbb{R}^{100}$\\ 
		$512\times4\times4$ conv. +BN+lReLU		& $512 \times4 \times 4$	\\
		$256\times4\times4$ conv. stride $1/2$+BN+lReLU		& $256 \times8 \times 8$ \\
		$128\times4\times4$ conv. stride $1/2$+BN+lReLU 		& $128 \times16 \times 16$ \\
		$64\times4\times4$ conv. 	stride $1/2$+BN+lReLU	& $64 \times32 \times 32$ \\
		$4\times4$ conv. stride $1/2$+BN+lReLU			& $3 \times64 \times 64$ \\\midrule
		$\tanh$			& $3 \times64 \times 64$ \\\bottomrule

\end{tabular}

\vspace{1cm}
\begin{tabular}{cc}
		\multicolumn{2}{c}{Discriminator 1 ($\text{D}_1$)}\\\toprule
		layer				&	Activation Size 	\\\midrule
		Input image							&	$3 \times64 \times 64$\\
		$64\times4\times4$ conv. stride $2$+BN+lReLU		& $64 \times32 \times 32$	\\
		$128\times4\times4$ conv. stride $2$+BN+lReLU		& $128 \times16 \times 16$ \\
		$256\times4\times4$ conv. stride $2$ +BN+lReLU		& $256 \times8 \times 8$ \\
		$512\times4\times4$ conv. stride $2$+BN+lReLU	& $512 \times4 \times 4$ \\
		$512\times4\times4$ conv.+BN+lReLU 			& $512 \times1 \times 1$ \\
		$2\times1\times1$ conv. +BN+lReLU			& $2 \times1 \times 1$ \\\midrule
		2-way softmax			& $2 \times1 \times 1$ \\\bottomrule

\end{tabular}

\vspace{1cm}
\begin{tabular}{cc}
		\multicolumn{2}{c}{Discriminator 2 ($\text{D}_2$)}\\\toprule
		layer				&	Activation Size 	\\\midrule
		Input image							&	$3 \times64 \times 64$\\
		$64\times4\times4$ conv. stride $2$			& $64 \times32 \times 32$	\\
		$128\times4\times4$ conv. stride $2$		& $128 \times16 \times 16$ \\
		$256\times4\times4$ conv. stride $2$		& $256 \times8 \times 8$ \\\midrule
		average								& $1$ \\\bottomrule		
\end{tabular}
\caption{Architecture of the proposed catWGAN. The feature representation was extracted from the third to the last layer of $\text{D}_1$.}
\label{tb:g}
\end{table}%

\section{Experiment setup}

\subsection{Datasets}
Two datasets were used in this research.
The first one is a fully annotated open access  dataset from the International Symposium on Biomedical Imaging (ISBI) 2016 Skin Lesion challenge. This dataset is part of the  International Skin Imaging Collaboration (ISIC) Archive, which is by far the largest publicly available dermoscopy image dataset. The 2016 challenge training dataset contains a total of 900 images with 173  melanomas and 727 benign cases. The test set is also released for analysis which consists of 75 melanomas and 304 benign cases. These are all 8-bit RGB colour images of varying spatial sizes. Segmentation masks were supplied by the organizer for the segmented lesion classification task. 

We applied the ground-truth segmentation mask  to extract the smallest square region that contains the lesion, and then resized it to  $256\times256$ with bilinear interpolation. The images in the training set were further augmented with rotation (in the range of [-\ang{180}, \ang{180}]), horizontal and vertical flipping, and elastic transform~\cite{simard2003best} to further boost the dataset. The size of the dataset for unsupervised training is 20k (balanced 10k+10k). As for the semi-supervised training, 70 images were randomly selected from each class of the original training set and augmented to a size of 10k (5k+5k). Translation and scale changes (in the range of [0.3, 1.5]) were used along with the previously mentioned augmentation techniques in the semi-supervised case  to alleviate overfitting. A python package named Augmentor~\cite{bloice2017augmentor} was used for the augmentation.


The second dataset is the $\text{PH}2$~\cite{barata2014two}.  It consists of a total of 200 dermoscopy images of melanocytic lesions, including 80 common nevi, 80 atypical nevi, and 40 melanomas.  Since in this work, we only solve the problem of binary classification, for this dataset, melanoma images constitute a class depicting malignancy and remaining images of common nevi and atypical nevi constitute another class depicting benign cases. The skin lesion was also extracted  using the method described above but without augmentation. The resultant images  served as the validation dataset to select the best model during the training process.

\subsection{Evaluation Metrics}
Sensitivity (SE), specificity (SP), accuracy (AC),  area under the receiver operating characteristic curve (AUC) and average precision (AP) are used for the evaluation as suggested by  the ISBI 2016 challenge~\cite{gutman2016skin}. They are defined mathematically as follows:
\begin{equation}
\begin{split}
\text{SE}&=\frac{\text{TP}}{\text{TP}+\text{FN}}\\
\text{SP}&=\frac{\text{TN}}{\text{TN}+\text{FP}}\\
\text{AC}&=\frac{\text{TP}+\text{TN}}{\text{TP}+\text{FP}+\text{FN}+\text{TN}}
\end{split}
\end{equation}
where $\text{TP}, \text{TN}, \text{FP}, \text{FN}$ represents the number of true positives, true negatives, false positives, false negatives respectively. AUC is defined as the integral of true positive rate (SE) with respect to the false positive rate (1-SP) under different thresholds. Similarly, AP is defined as the integral of precision with respect to recall under different thresholds. Precision and recall are expressed mathematically as:
\begin{equation}
\begin{split}
\text{Precision}=\frac{\text{T}P}{\text{TP}+\text{FP}}\\
\text{Recall} = \frac{\text{TP}}{\text{TP}+\text{FN}}\\
\end{split}
\end{equation}

AP is used as the ranking metric for different methods because it is more sensitive to the change of TP. Note that there are different variants of  the implementation of AP. To be consistent with the other works that also use the ISBI challenge dataset, we use the ``scikit-learn'' Python package for the computation of all the aforementioned metrics.

\subsection{Implementation details}
All the networks were trained on the Guillimin cluster of Calcul Qu\'ebec.  Adam optimizer~\cite{kingma2014adam} with $\beta_1=0.5$ and $\beta_2=0.9$ was used for all three networks with learning rate  0.0002. Batch size was chosen to be 200 to get a good estimate of the marginal entropy of the real and generated sample. $\text{D}_2$ was trained 5 times more often than $\text{G}$ and $\text{D}_1$ to ensure the 1-Lipschitz assumption. $\lambda$ and $\alpha$ were set to be 10 and 0.1. The implementation was based on the PyTorch framework.  Training was stopped after 16K iterations. 

%

\subsection{Experiments}

Three experiments were conducted to show the effectiveness of the proposed method. First, we monitored how the quality of our features evolves during the training of the catWGAN.  We sampled the network every 50 iterations and extracted features from the third to last layer of $\text{D}_1$ and trained a linear SVM on top. PH2 dataset   served as the validation dataset and 5-fold cross-validation was performed.  
Second, we performed horizontal comparison to the aforementioned baseline methods on the 2016 ISIC challenge test dataset.  
Lastly, we evaluated the  image generated from the trained generator.

\pgfplotsset{tick scale binop=\times}

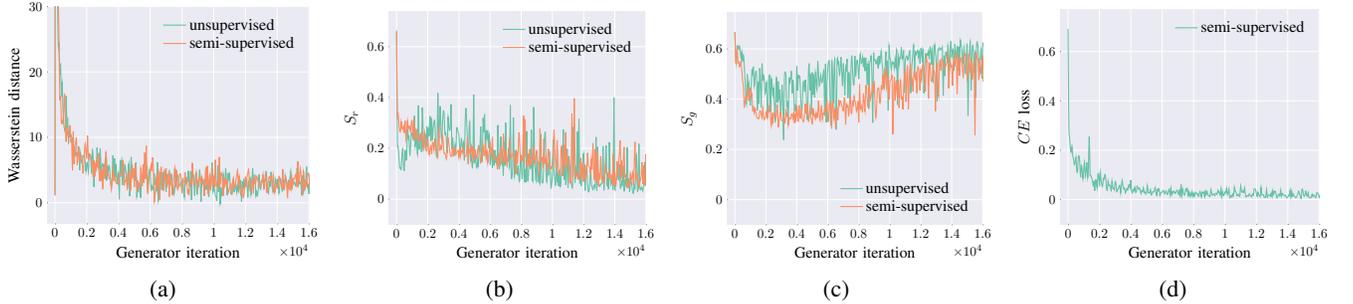
\begin{figure*}[htbp]
 \centering 
\begin{subfigure}[b]{0.24\textwidth}
\centering
\resizebox{\textwidth}{!}{
\begin{tikzpicture}
\begin{axis}[xlabel= \large Generator iteration,ylabel=\large Wasserstein distance, xmin=-500,xmax=16000, ymax=30,
ylabel near ticks,xlabel near ticks,mark size=0.7pt,
tick pos=left,tickwidth=1mm,legend pos= north east, legend entries={unsupervised, semi-supervised},axisStyle,legend style={font=\large}]
\addplot[ smooth,Set2-A,thick] plot coordinates {
(1,1.11835)(51,186.74700)(101,79.35310)(151,38.18790)(201,24.43960)(251,24.35980)(301,19.56240)(351,21.19820)(401,16.41900)(451,18.36860)(501,14.30010)(551,11.94820)(601,11.38960)(651,11.16650)(701,16.36350)(751,11.49290)(801,12.83280)(851,9.14293)(901,10.85200)(951,10.08460)(1001,10.39670)(1051,9.99171)(1101,9.66569)(1151,7.03338)(1201,8.41039)(1251,8.42722)(1301,7.14746)(1351,7.65481)(1401,7.45842)(1451,7.33479)(1501,8.50845)(1551,4.97502)(1601,7.59052)(1651,6.46500)(1701,6.07104)(1751,7.05389)(1801,7.12202)(1851,7.25666)(1901,6.06194)(1951,4.80565)(2001,6.12375)(2051,5.83090)(2101,4.70386)(2151,7.26888)(2201,7.18554)(2251,7.05608)(2301,4.47525)(2351,8.01494)(2401,7.01751)(2451,2.43489)(2501,5.72610)(2551,5.80411)(2601,5.12690)(2651,8.44778)(2701,5.51474)(2751,7.34542)(2801,3.34841)(2851,8.30018)(2901,4.80308)(2951,5.44879)(3001,5.95625)(3051,5.47920)(3101,4.24120)(3151,6.19044)(3201,6.88953)(3251,3.39228)(3301,5.53670)(3351,4.54745)(3401,5.65842)(3451,4.99120)(3501,3.26644)(3551,2.40736)(3601,3.87354)(3651,7.22192)(3701,4.04284)(3751,6.76519)(3801,3.93382)(3851,5.92825)(3901,3.01749)(3951,3.79743)(4001,4.48375)(4051,0.53998)(4101,6.37645)(4151,4.41066)(4201,6.51969)(4251,2.70210)(4301,5.93331)(4351,3.46429)(4401,5.73385)(4451,4.99689)(4501,4.40658)(4551,2.37575)(4601,4.39143)(4651,5.94178)(4701,0.99992)(4751,3.51938)(4801,6.99070)(4851,3.76279)(4901,4.96482)(4951,5.18548)(5001,5.03375)(5051,3.39787)(5101,6.19430)(5151,0.87832)(5201,3.38946)(5251,1.59534)(5301,2.87363)(5351,2.44800)(5401,0.75241)(5451,3.78603)(5501,4.18259)(5551,4.03469)(5601,5.21581)(5651,6.22249)(5701,3.67893)(5751,5.16059)(5801,5.48174)(5851,2.71249)(5901,3.81319)(5951,5.07735)(6001,4.58939)(6051,1.88452)(6101,2.58418)(6151,2.17091)(6201,2.10584)(6251,4.49188)(6301,2.90417)(6351,2.87289)(6401,1.90681)(6451,6.19542)(6501,0.44231)(6551,1.93626)(6601,4.06110)(6651,4.58558)(6701,2.21397)(6751,2.28083)(6801,4.62102)(6851,2.66519)(6901,4.79691)(6951,3.48181)(7001,1.65056)(7051,4.19006)(7101,3.99522)(7151,4.34004)(7201,4.87242)(7251,3.81736)(7301,3.66609)(7351,4.34080)(7401,2.25266)(7451,2.64270)(7501,1.37643)(7551,2.42901)(7601,4.99511)(7651,4.74714)(7701,4.34071)(7751,3.98794)(7801,2.56998)(7851,2.63322)(7901,2.87749)(7951,3.26751)(8001,1.69588)(8051,3.78342)(8101,2.07582)(8151,0.84192)(8201,3.72599)(8251,2.99094)(8301,3.41494)(8351,3.26186)(8401,4.88186)(8451,2.86792)(8501,2.46860)(8551,4.79368)(8601,0.41767)(8651,5.02633)(8701,4.35033)(8751,3.85510)(8801,1.10222)(8851,4.08875)(8901,4.34479)(8951,1.25006)(9001,3.30312)(9051,2.46009)(9101,4.96967)(9151,3.80059)(9201,0.17662)(9251,3.89102)(9301,2.92911)(9351,3.43811)(9401,1.48119)(9451,3.76256)(9501,2.22530)(9551,2.14010)(9601,3.82470)(9651,1.63936)(9701,1.06898)(9751,4.24279)(9801,2.11403)(9851,4.08147)(9901,6.12030)(9951,1.82967)(10001,2.61896)(10051,2.00253)(10101,1.11323)(10151,0.93479)(10201,0.56894)(10251,5.65795)(10301,1.67998)(10351,5.01885)(10401,-0.20397)(10451,2.25324)(10501,1.21194)(10551,2.49269)(10601,4.40610)(10651,3.39830)(10701,2.73784)(10751,5.28690)(10801,3.48628)(10851,1.13755)(10901,4.39816)(10951,3.60928)(11001,4.56159)(11051,1.69443)(11101,1.34269)(11151,1.55572)(11201,3.17809)(11251,1.93957)(11301,1.66713)(11351,2.13871)(11401,2.63121)(11451,1.58894)(11501,2.80713)(11551,1.75059)(11601,3.29206)(11651,2.34786)(11701,3.35883)(11751,0.13280)(11801,3.21422)(11851,2.17966)(11901,3.66955)(11951,2.35353)(12001,2.76516)(12051,3.52417)(12101,3.06701)(12151,0.82652)(12201,3.18879)(12251,3.28517)(12301,3.73000)(12351,1.17594)(12401,2.39658)(12451,5.29370)(12501,3.09257)(12551,3.38612)(12601,4.18188)(12651,1.07878)(12701,1.71305)(12751,1.13118)(12801,1.95333)(12851,4.32752)(12901,2.62148)(12951,2.05916)(13001,2.81423)(13051,3.23824)(13101,3.80882)(13151,2.50708)(13201,3.47248)(13251,3.43631)(13301,4.84067)(13351,3.26141)(13401,2.17308)(13451,1.45293)(13501,3.24813)(13551,2.49038)(13601,4.60011)(13651,4.12788)(13701,2.41236)(13751,1.69367)(13801,2.64176)(13851,2.11541)(13901,3.62874)(13951,2.86072)(14001,2.35482)(14051,3.64476)(14101,1.43792)(14151,2.35173)(14201,3.69926)(14251,3.13749)(14301,1.81371)(14351,2.05045)(14401,3.46027)(14451,2.03406)(14501,4.18301)(14551,3.33505)(14601,3.20592)(14651,1.87521)(14701,3.35972)(14751,3.16537)(14801,3.56923)(14851,4.76116)(14901,2.21955)(14951,3.45031)(15001,2.84571)(15051,2.88278)(15101,3.15364)(15151,2.55596)(15201,2.93473)(15251,5.31127)(15301,2.54915)(15351,3.33799)(15401,2.87765)(15451,3.18779)(15501,2.68148)(15551,1.13805)(15601,4.20232)(15651,1.56917)(15701,2.53791)(15751,3.36272)(15801,3.40762)(15851,5.50732)(15901,2.52333)(15951,2.84061)(16001,1.34014)(16051,3.87734)(16101,2.40969)(16151,4.27056)(16201,3.23073)(16251,2.68008)
};
\addplot[ smooth,Set2-B,thick] plot coordinates {
(1,1.14402)(51,145.28900)(101,54.19190)(151,37.84230)(201,32.27590)(251,24.86800)(301,16.06930)(351,16.21170)(401,12.22470)(451,12.53410)(501,12.26590)(551,14.44020)(601,16.69300)(651,11.88380)(701,9.32601)(751,11.45140)(801,10.62080)(851,9.74653)(901,11.24260)(951,11.70940)(1001,9.33604)(1051,9.50479)(1101,4.95822)(1151,10.86070)(1201,5.92242)(1251,8.32547)(1301,9.25093)(1351,9.76751)(1401,5.76620)(1451,7.28865)(1501,8.15283)(1551,6.42525)(1601,6.91654)(1651,7.22356)(1701,6.87613)(1751,4.59141)(1801,5.11385)(1851,7.81450)(1901,8.97919)(1951,7.54052)(2001,6.95029)(2051,10.24230)(2101,6.79610)(2151,7.50383)(2201,7.29679)(2251,7.06726)(2301,6.00552)(2351,6.12525)(2401,3.92937)(2451,6.50304)(2501,3.99171)(2551,6.07179)(2601,5.25273)(2651,6.57015)(2701,5.41023)(2751,7.55306)(2801,3.44681)(2851,4.44728)(2901,3.29767)(2951,5.17250)(3001,2.06808)(3051,4.80433)(3101,5.81975)(3151,3.65424)(3201,7.61439)(3251,6.57117)(3301,4.20913)(3351,4.61920)(3401,3.70414)(3451,3.71556)(3501,3.40004)(3551,2.03706)(3601,3.43367)(3651,5.45923)(3701,3.35482)(3751,4.13161)(3801,3.40070)(3851,3.06669)(3901,2.90115)(3951,3.11205)(4001,2.85349)(4051,4.02734)(4101,4.49475)(4151,4.06797)(4201,4.25788)(4251,6.37963)(4301,5.91746)(4351,2.83063)(4401,3.29754)(4451,3.34291)(4501,4.20502)(4551,2.12972)(4601,2.93578)(4651,4.18523)(4701,4.78417)(4751,1.84487)(4801,3.41506)(4851,5.90768)(4901,3.91766)(4951,4.90583)(5001,2.42332)(5051,5.97777)(5101,4.46296)(5151,3.25601)(5201,1.66838)(5251,3.08216)(5301,2.51602)(5351,2.54791)(5401,5.32871)(5451,4.14707)(5501,2.54194)(5551,5.80402)(5601,3.03083)(5651,7.65879)(5701,3.66122)(5751,8.69145)(5801,2.60095)(5851,1.45650)(5901,3.23370)(5951,3.90966)(6001,4.74484)(6051,5.87003)(6101,4.92972)(6151,6.95027)(6201,5.00964)(6251,0.24972)(6301,1.30499)(6351,2.45290)(6401,3.93607)(6451,2.37203)(6501,4.94844)(6551,3.40709)(6601,0.48554)(6651,4.58463)(6701,5.37311)(6751,3.16297)(6801,2.94582)(6851,1.87774)(6901,4.87887)(6951,3.59898)(7001,5.28424)(7051,2.53735)(7101,3.10666)(7151,4.22582)(7201,0.94698)(7251,4.50898)(7301,2.79341)(7351,2.61905)(7401,1.13417)(7451,3.18001)(7501,4.17519)(7551,5.32347)(7601,2.35158)(7651,4.93920)(7701,2.31170)(7751,2.05274)(7801,3.19942)(7851,4.29042)(7901,4.37582)(7951,3.59749)(8001,1.43454)(8051,3.11552)(8101,3.18559)(8151,3.37048)(8201,4.62429)(8251,2.95716)(8301,3.80430)(8351,1.54385)(8401,1.98400)(8451,1.17756)(8501,3.20251)(8551,3.52932)(8601,2.93323)(8651,6.08992)(8701,3.47898)(8751,2.68942)(8801,2.98610)(8851,3.30472)(8901,4.59305)(8951,3.18547)(9001,2.17569)(9051,3.42581)(9101,3.92620)(9151,3.99596)(9201,3.04176)(9251,3.88675)(9301,3.08739)(9351,3.34494)(9401,1.04424)(9451,3.16091)(9501,5.37274)(9551,2.58676)(9601,5.63876)(9651,1.38715)(9701,3.28104)(9751,4.26258)(9801,1.88680)(9851,3.34092)(9901,5.32642)(9951,2.95081)(10001,3.94869)(10051,2.46061)(10101,6.57136)(10151,2.70992)(10201,3.01036)(10251,2.13654)(10301,2.49734)(10351,6.98375)(10401,2.96605)(10451,4.36524)(10501,4.54583)(10551,3.25103)(10601,1.03921)(10651,3.52894)(10701,2.55220)(10751,3.72142)(10801,2.88412)(10851,3.41854)(10901,1.60918)(10951,1.95348)(11001,2.21851)(11051,2.80632)(11101,2.81818)(11151,1.25862)(11201,2.90858)(11251,3.80872)(11301,2.26444)(11351,2.71174)(11401,3.48978)(11451,4.18864)(11501,4.98990)(11551,2.79390)(11601,4.06484)(11651,2.15061)(11701,1.29767)(11751,2.06240)(11801,2.56002)(11851,0.81691)(11901,1.11828)(11951,2.21932)(12001,3.97525)(12051,2.49881)(12101,4.12226)(12151,4.53906)(12201,5.05595)(12251,2.81451)(12301,1.80476)(12351,2.64668)(12401,3.57832)(12451,2.99984)(12501,1.43353)(12551,4.12830)(12601,2.95012)(12651,3.08307)(12701,4.03424)(12751,3.66369)(12801,3.26158)(12851,2.61350)(12901,2.38380)(12951,2.66155)(13001,3.42764)(13051,3.29830)(13101,4.30003)(13151,2.23063)(13201,2.46486)(13251,4.35475)(13301,3.78349)(13351,1.50023)(13401,2.79108)(13451,3.50506)(13501,3.98856)(13551,3.67440)(13601,4.10538)(13651,3.26591)(13701,2.02818)(13751,4.22226)(13801,2.92305)(13851,1.53951)(13901,2.37161)(13951,2.40379)(14001,2.78853)(14051,2.24997)(14101,3.97076)(14151,3.58773)(14201,2.92236)(14251,3.01140)(14301,2.43221)(14351,5.12514)(14401,2.17446)(14451,2.37072)(14501,3.75972)(14551,3.83005)(14601,2.63827)(14651,0.70887)(14701,3.05612)(14751,2.21777)(14801,3.18131)(14851,1.94948)(14901,4.34014)(14951,2.32946)(15001,3.19578)(15051,3.80371)(15101,3.61280)(15151,4.29676)(15201,4.17741)(15251,2.37283)(15301,6.31655)(15351,3.11378)(15401,3.82091)(15451,5.22804)(15501,4.07002)(15551,2.23012)(15601,1.38994)(15651,4.79989)(15701,3.54237)(15751,1.92726)(15801,3.22498)(15851,2.52830)(15901,4.00703)(15951,4.56269)(16001,4.15310)(16051,2.28690)(16101,4.59285)(16151,3.71957)(16201,4.59439)(16251,2.58422)
};
\end{axis}
\end{tikzpicture}
}
\caption*{(a)}
\end{subfigure}
\begin{subfigure}[b]{0.24\textwidth}
\centering
\resizebox{\textwidth}{!}{
\begin{tikzpicture}
\begin{axis}[xlabel=\large Generator iteration,ylabel=\large $S_r$, xmin=-500,xmax=16000,ymin=-0.1,
ylabel near ticks,xlabel near ticks,
tick pos=left,tickwidth=1mm, legend pos= north east, legend entries={unsupervised, semi-supervised},axisStyle,legend style={font=\large}]
\addplot[ smooth,Set2-A,thick] plot coordinates {
(1,0.65695)(51,0.24293)(101,0.20438)(151,0.16687)(201,0.12430)(251,0.12380)(301,0.11794)(351,0.17401)(401,0.19967)(451,0.12980)(501,0.11477)(551,0.19304)(601,0.15635)(651,0.16485)(701,0.29260)(751,0.23923)(801,0.24400)(851,0.17146)(901,0.20871)(951,0.25844)(1001,0.25342)(1051,0.24043)(1101,0.23809)(1151,0.32729)(1201,0.20677)(1251,0.26764)(1301,0.22141)(1351,0.18300)(1401,0.36876)(1451,0.20011)(1501,0.23927)(1551,0.25480)(1601,0.18701)(1651,0.35995)(1701,0.27908)(1751,0.19757)(1801,0.22509)(1851,0.25688)(1901,0.20578)(1951,0.28465)(2001,0.22928)(2051,0.27623)(2101,0.24942)(2151,0.32644)(2201,0.22820)(2251,0.31855)(2301,0.23362)(2351,0.27963)(2401,0.30305)(2451,0.30687)(2501,0.23753)(2551,0.17292)(2601,0.19303)(2651,0.41721)(2701,0.23953)(2751,0.28077)(2801,0.36569)(2851,0.27099)(2901,0.23435)(2951,0.30204)(3001,0.34572)(3051,0.33479)(3101,0.25816)(3151,0.26144)(3201,0.23034)(3251,0.18159)(3301,0.31906)(3351,0.28810)(3401,0.27002)(3451,0.28535)(3501,0.37315)(3551,0.19965)(3601,0.24325)(3651,0.18591)(3701,0.17273)(3751,0.21245)(3801,0.17950)(3851,0.27313)(3901,0.27948)(3951,0.26225)(4001,0.25270)(4051,0.20444)(4101,0.18628)(4151,0.23678)(4201,0.21903)(4251,0.18640)(4301,0.22424)(4351,0.15216)(4401,0.29986)(4451,0.35784)(4501,0.30004)(4551,0.23341)(4601,0.14290)(4651,0.26183)(4701,0.29948)(4751,0.15657)(4801,0.17271)(4851,0.12608)(4901,0.15588)(4951,0.41163)(5001,0.15410)(5051,0.17282)(5101,0.14130)(5151,0.12395)(5201,0.15571)(5251,0.12706)(5301,0.20567)(5351,0.18994)(5401,0.28485)(5451,0.16309)(5501,0.16116)(5551,0.31827)(5601,0.26305)(5651,0.33021)(5701,0.16949)(5751,0.26580)(5801,0.24022)(5851,0.26829)(5901,0.20195)(5951,0.25462)(6001,0.14148)(6051,0.16251)(6101,0.20195)(6151,0.12215)(6201,0.19391)(6251,0.13213)(6301,0.19271)(6351,0.14447)(6401,0.16443)(6451,0.14164)(6501,0.29112)(6551,0.18055)(6601,0.11943)(6651,0.20509)(6701,0.24333)(6751,0.13960)(6801,0.16409)(6851,0.09134)(6901,0.22742)(6951,0.13214)(7001,0.30110)(7051,0.16460)(7101,0.15339)(7151,0.15375)(7201,0.20203)(7251,0.12543)(7301,0.36965)(7351,0.10304)(7401,0.08588)(7451,0.13322)(7501,0.32131)(7551,0.09580)(7601,0.13226)(7651,0.19247)(7701,0.11844)(7751,0.09562)(7801,0.13359)(7851,0.07986)(7901,0.07629)(7951,0.17010)(8001,0.13927)(8051,0.08036)(8101,0.12388)(8151,0.31746)(8201,0.07392)(8251,0.11821)(8301,0.07014)(8351,0.26720)(8401,0.08353)(8451,0.08610)(8501,0.14110)(8551,0.17588)(8601,0.12025)(8651,0.27267)(8701,0.09362)(8751,0.09261)(8801,0.36191)(8851,0.17129)(8901,0.11146)(8951,0.07382)(9001,0.18514)(9051,0.07743)(9101,0.07681)(9151,0.14114)(9201,0.06024)(9251,0.25504)(9301,0.08389)(9351,0.06493)(9401,0.07189)(9451,0.14655)(9501,0.24592)(9551,0.25402)(9601,0.10022)(9651,0.06784)(9701,0.13421)(9751,0.09672)(9801,0.18748)(9851,0.13276)(9901,0.04637)(9951,0.05847)(10001,0.10862)(10051,0.23438)(10101,0.10692)(10151,0.08230)(10201,0.05669)(10251,0.15569)(10301,0.06748)(10351,0.09890)(10401,0.05101)(10451,0.11504)(10501,0.05062)(10551,0.06853)(10601,0.12800)(10651,0.13494)(10701,0.27344)(10751,0.07624)(10801,0.17036)(10851,0.05638)(10901,0.10272)(10951,0.19755)(11001,0.04428)(11051,0.16194)(11101,0.14844)(11151,0.08472)(11201,0.07023)(11251,0.04565)(11301,0.06279)(11351,0.28470)(11401,0.13895)(11451,0.06501)(11501,0.13006)(11551,0.09465)(11601,0.26533)(11651,0.07922)(11701,0.31501)(11751,0.07210)(11801,0.08787)(11851,0.05203)(11901,0.11370)(11951,0.15053)(12001,0.17930)(12051,0.10746)(12101,0.03376)(12151,0.10507)(12201,0.13248)(12251,0.07020)(12301,0.03089)(12351,0.03375)(12401,0.06296)(12451,0.11694)(12501,0.06019)(12551,0.04590)(12601,0.03707)(12651,0.05368)(12701,0.05553)(12751,0.05454)(12801,0.24408)(12851,0.03291)(12901,0.20550)(12951,0.07183)(13001,0.05747)(13051,0.05058)(13101,0.05949)(13151,0.06084)(13201,0.08193)(13251,0.20871)(13301,0.03879)(13351,0.08186)(13401,0.14612)(13451,0.21053)(13501,0.05455)(13551,0.10330)(13601,0.04088)(13651,0.22382)(13701,0.09627)(13751,0.05741)(13801,0.08228)(13851,0.07595)(13901,0.05297)(13951,0.40015)(14001,0.05140)(14051,0.07150)(14101,0.07584)(14151,0.09439)(14201,0.08703)(14251,0.07814)(14301,0.13648)(14351,0.04824)(14401,0.05031)(14451,0.07006)(14501,0.05697)(14551,0.13075)(14601,0.03059)(14651,0.07233)(14701,0.07941)(14751,0.05732)(14801,0.20374)(14851,0.04898)(14901,0.08238)(14951,0.02867)(15001,0.06341)(15051,0.07210)(15101,0.05214)(15151,0.02685)(15201,0.02479)(15251,0.05652)(15301,0.04416)(15351,0.13618)(15401,0.16970)(15451,0.03828)(15501,0.05322)(15551,0.21766)(15601,0.05707)(15651,0.03615)(15701,0.04730)(15751,0.04247)(15801,0.03148)(15851,0.05885)(15901,0.08935)(15951,0.14625)(16001,0.05894)(16051,0.07833)(16101,0.13689)(16151,0.16680)(16201,0.14598)(16251,0.06978)
};
\addplot[ smooth,Set2-B,thick] plot coordinates {
(1,0.66365)(51,0.33943)(101,0.33468)(151,0.30268)(201,0.26840)(251,0.26694)(301,0.30525)(351,0.28021)(401,0.27425)(451,0.28714)(501,0.26021)(551,0.30779)(601,0.23523)(651,0.26814)(701,0.30769)(751,0.27664)(801,0.28719)(851,0.28377)(901,0.26572)(951,0.30078)(1001,0.30640)(1051,0.24175)(1101,0.31031)(1151,0.27725)(1201,0.23362)(1251,0.27448)(1301,0.26387)(1351,0.23609)(1401,0.24321)(1451,0.22912)(1501,0.24660)(1551,0.23654)(1601,0.24753)(1651,0.26485)(1701,0.20233)(1751,0.23394)(1801,0.26267)(1851,0.20879)(1901,0.19168)(1951,0.19310)(2001,0.23996)(2051,0.19794)(2101,0.17467)(2151,0.19860)(2201,0.21039)(2251,0.17801)(2301,0.22045)(2351,0.21936)(2401,0.19358)(2451,0.23834)(2501,0.19851)(2551,0.19172)(2601,0.18993)(2651,0.16784)(2701,0.29100)(2751,0.20044)(2801,0.19298)(2851,0.20957)(2901,0.19693)(2951,0.25297)(3001,0.21234)(3051,0.18139)(3101,0.18005)(3151,0.17223)(3201,0.15781)(3251,0.18994)(3301,0.24091)(3351,0.22088)(3401,0.19314)(3451,0.17440)(3501,0.16756)(3551,0.17276)(3601,0.20853)(3651,0.16772)(3701,0.19982)(3751,0.19778)(3801,0.17565)(3851,0.18439)(3901,0.18656)(3951,0.19050)(4001,0.16396)(4051,0.20347)(4101,0.16780)(4151,0.19092)(4201,0.18385)(4251,0.20345)(4301,0.18538)(4351,0.17292)(4401,0.17033)(4451,0.16918)(4501,0.19102)(4551,0.24409)(4601,0.13668)(4651,0.19493)(4701,0.18313)(4751,0.18767)(4801,0.20870)(4851,0.16433)(4901,0.20723)(4951,0.15400)(5001,0.18005)(5051,0.19236)(5101,0.16890)(5151,0.16052)(5201,0.18776)(5251,0.20737)(5301,0.18584)(5351,0.18567)(5401,0.14362)(5451,0.15967)(5501,0.18478)(5551,0.20917)(5601,0.17260)(5651,0.18839)(5701,0.21811)(5751,0.15822)(5801,0.18988)(5851,0.14271)(5901,0.17141)(5951,0.19106)(6001,0.21973)(6051,0.18523)(6101,0.18148)(6151,0.24416)(6201,0.15401)(6251,0.17613)(6301,0.16400)(6351,0.16037)(6401,0.16327)(6451,0.12722)(6501,0.12253)(6551,0.18142)(6601,0.14497)(6651,0.13402)(6701,0.19948)(6751,0.19368)(6801,0.14254)(6851,0.25766)(6901,0.20300)(6951,0.14772)(7001,0.18934)(7051,0.14634)(7101,0.12957)(7151,0.17675)(7201,0.19028)(7251,0.23984)(7301,0.16211)(7351,0.17186)(7401,0.17164)(7451,0.17196)(7501,0.18484)(7551,0.15032)(7601,0.16032)(7651,0.09916)(7701,0.18632)(7751,0.13829)(7801,0.18769)(7851,0.26178)(7901,0.22025)(7951,0.14982)(8001,0.16974)(8051,0.15901)(8101,0.16176)(8151,0.32876)(8201,0.16540)(8251,0.31393)(8301,0.14792)(8351,0.14215)(8401,0.12895)(8451,0.17898)(8501,0.12894)(8551,0.19743)(8601,0.13641)(8651,0.13975)(8701,0.16474)(8751,0.15677)(8801,0.20784)(8851,0.14102)(8901,0.16279)(8951,0.18033)(9001,0.18941)(9051,0.14996)(9101,0.14056)(9151,0.10766)(9201,0.27629)(9251,0.19709)(9301,0.14671)(9351,0.16207)(9401,0.12644)(9451,0.23690)(9501,0.14312)(9551,0.15197)(9601,0.16440)(9651,0.11215)(9701,0.17201)(9751,0.19540)(9801,0.13972)(9851,0.11955)(9901,0.12698)(9951,0.12104)(10001,0.14556)(10051,0.18042)(10101,0.27329)(10151,0.19700)(10201,0.14685)(10251,0.12165)(10301,0.23203)(10351,0.15147)(10401,0.09795)(10451,0.10753)(10501,0.10922)(10551,0.18315)(10601,0.20494)(10651,0.09053)(10701,0.14958)(10751,0.21287)(10801,0.13015)(10851,0.26264)(10901,0.12468)(10951,0.17114)(11001,0.15419)(11051,0.14390)(11101,0.06765)(11151,0.07634)(11201,0.07940)(11251,0.33622)(11301,0.09751)(11351,0.13810)(11401,0.39634)(11451,0.15808)(11501,0.08350)(11551,0.10682)(11601,0.09366)(11651,0.07754)(11701,0.08629)(11751,0.12053)(11801,0.16224)(11851,0.07931)(11901,0.14134)(11951,0.11996)(12001,0.06883)(12051,0.13625)(12101,0.08208)(12151,0.08355)(12201,0.10185)(12251,0.18178)(12301,0.13498)(12351,0.12881)(12401,0.12372)(12451,0.10283)(12501,0.15838)(12551,0.11916)(12601,0.22015)(12651,0.13287)(12701,0.07443)(12751,0.19739)(12801,0.07310)(12851,0.09135)(12901,0.18560)(12951,0.05285)(13001,0.09909)(13051,0.08019)(13101,0.16790)(13151,0.08258)(13201,0.09095)(13251,0.06714)(13301,0.06140)(13351,0.08110)(13401,0.13329)(13451,0.10340)(13501,0.09950)(13551,0.08269)(13601,0.08678)(13651,0.10915)(13701,0.08257)(13751,0.09483)(13801,0.09237)(13851,0.06260)(13901,0.17730)(13951,0.05794)(14001,0.10864)(14051,0.05269)(14101,0.06629)(14151,0.14836)(14201,0.08999)(14251,0.14430)(14301,0.18182)(14351,0.23027)(14401,0.08079)(14451,0.15119)(14501,0.09077)(14551,0.14865)(14601,0.20501)(14651,0.09092)(14701,0.09451)(14751,0.04962)(14801,0.05389)(14851,0.06699)(14901,0.04688)(14951,0.05453)(15001,0.09877)(15051,0.07517)(15101,0.08602)(15151,0.10582)(15201,0.05608)(15251,0.14449)(15301,0.22498)(15351,0.07774)(15401,0.10248)(15451,0.08679)(15501,0.07397)(15551,0.04652)(15601,0.06563)(15651,0.07727)(15701,0.09084)(15751,0.14440)(15801,0.05957)(15851,0.13153)(15901,0.07286)(15951,0.05945)(16001,0.17611)(16051,0.06997)(16101,0.06001)(16151,0.04867)(16201,0.07591)(16251,0.10532)
};
\end{axis}
\end{tikzpicture}
}
\caption*{(b)}
\end{subfigure}
\begin{subfigure}[b]{0.24\textwidth}
\centering
\resizebox{\textwidth}{!}{
\begin{tikzpicture}
\begin{axis}[xlabel=\large Generator iteration,ylabel=\large $S_g$, xmin=-500,xmax=16000,ymin=-0.1,
ylabel near ticks,xlabel near ticks,
tick pos=left,tickwidth=1mm,legend pos= south east, legend entries={unsupervised, semi-supervised},axisStyle,legend style={font=\large}]
\addplot[ smooth,Set2-A,thick] plot coordinates {
(1,0.66558)(51,0.58193)(101,0.54085)(151,0.60738)(201,0.61022)(251,0.59890)(301,0.61132)(351,0.58772)(401,0.56276)(451,0.58792)(501,0.55145)(551,0.57459)(601,0.54247)(651,0.47754)(701,0.42166)(751,0.38087)(801,0.49899)(851,0.42562)(901,0.46526)(951,0.41754)(1001,0.37445)(1051,0.43516)(1101,0.51679)(1151,0.49817)(1201,0.52792)(1251,0.45243)(1301,0.48250)(1351,0.50846)(1401,0.49820)(1451,0.35567)(1501,0.47029)(1551,0.38134)(1601,0.52260)(1651,0.46206)(1701,0.36605)(1751,0.54026)(1801,0.47502)(1851,0.49464)(1901,0.47552)(1951,0.42145)(2001,0.44018)(2051,0.46270)(2101,0.32378)(2151,0.38258)(2201,0.48411)(2251,0.48409)(2301,0.34829)(2351,0.47810)(2401,0.44901)(2451,0.41451)(2501,0.44531)(2551,0.39842)(2601,0.30689)(2651,0.44155)(2701,0.48468)(2751,0.38642)(2801,0.47062)(2851,0.40289)(2901,0.43654)(2951,0.33786)(3001,0.38786)(3051,0.45624)(3101,0.37755)(3151,0.23783)(3201,0.48002)(3251,0.50491)(3301,0.48951)(3351,0.37170)(3401,0.38988)(3451,0.30806)(3501,0.47822)(3551,0.46543)(3601,0.48623)(3651,0.50916)(3701,0.42240)(3751,0.47553)(3801,0.42506)(3851,0.35663)(3901,0.37273)(3951,0.35824)(4001,0.53084)(4051,0.43412)(4101,0.49302)(4151,0.48614)(4201,0.48577)(4251,0.40746)(4301,0.35640)(4351,0.37071)(4401,0.45873)(4451,0.46848)(4501,0.42983)(4551,0.47696)(4601,0.43273)(4651,0.33766)(4701,0.32881)(4751,0.50189)(4801,0.53687)(4851,0.43583)(4901,0.49723)(4951,0.34045)(5001,0.50641)(5051,0.41736)(5101,0.49843)(5151,0.40679)(5201,0.50660)(5251,0.47126)(5301,0.43691)(5351,0.55138)(5401,0.50376)(5451,0.45341)(5501,0.47287)(5551,0.49674)(5601,0.46744)(5651,0.47042)(5701,0.49788)(5751,0.50536)(5801,0.33428)(5851,0.54012)(5901,0.30754)(5951,0.45821)(6001,0.52406)(6051,0.54997)(6101,0.54040)(6151,0.37942)(6201,0.30497)(6251,0.53775)(6301,0.51799)(6351,0.38412)(6401,0.50187)(6451,0.55067)(6501,0.31868)(6551,0.29710)(6601,0.54931)(6651,0.54174)(6701,0.54851)(6751,0.49688)(6801,0.51647)(6851,0.56711)(6901,0.51106)(6951,0.45269)(7001,0.53991)(7051,0.50256)(7101,0.41602)(7151,0.52716)(7201,0.51406)(7251,0.55678)(7301,0.40432)(7351,0.50098)(7401,0.48726)(7451,0.39176)(7501,0.51992)(7551,0.57318)(7601,0.50908)(7651,0.46051)(7701,0.46566)(7751,0.57137)(7801,0.49144)(7851,0.54087)(7901,0.55510)(7951,0.53328)(8001,0.48036)(8051,0.55276)(8101,0.58005)(8151,0.54283)(8201,0.54678)(8251,0.56889)(8301,0.52779)(8351,0.56505)(8401,0.51145)(8451,0.56491)(8501,0.51499)(8551,0.45687)(8601,0.54114)(8651,0.36217)(8701,0.53408)(8751,0.55324)(8801,0.56954)(8851,0.48770)(8901,0.30614)(8951,0.57356)(9001,0.55369)(9051,0.55774)(9101,0.54100)(9151,0.54390)(9201,0.55973)(9251,0.52259)(9301,0.57225)(9351,0.54269)(9401,0.56327)(9451,0.59117)(9501,0.54337)(9551,0.51486)(9601,0.44186)(9651,0.50596)(9701,0.53039)(9751,0.58887)(9801,0.48354)(9851,0.55211)(9901,0.35462)(9951,0.60559)(10001,0.56818)(10051,0.55412)(10101,0.51703)(10151,0.54754)(10201,0.54090)(10251,0.54496)(10301,0.52235)(10351,0.56342)(10401,0.52586)(10451,0.59474)(10501,0.37749)(10551,0.56579)(10601,0.53651)(10651,0.56379)(10701,0.56525)(10751,0.60629)(10801,0.55993)(10851,0.56580)(10901,0.58665)(10951,0.42519)(11001,0.54758)(11051,0.56541)(11101,0.57211)(11151,0.55566)(11201,0.56409)(11251,0.58203)(11301,0.57873)(11351,0.42371)(11401,0.50465)(11451,0.54935)(11501,0.58994)(11551,0.61312)(11601,0.51035)(11651,0.58609)(11701,0.57600)(11751,0.58751)(11801,0.59413)(11851,0.56682)(11901,0.59064)(11951,0.54218)(12001,0.59979)(12051,0.58944)(12101,0.54639)(12151,0.59633)(12201,0.42342)(12251,0.52882)(12301,0.57897)(12351,0.57386)(12401,0.58105)(12451,0.57432)(12501,0.59121)(12551,0.55869)(12601,0.60567)(12651,0.58935)(12701,0.53662)(12751,0.60132)(12801,0.46496)(12851,0.57720)(12901,0.61615)(12951,0.57846)(13001,0.54548)(13051,0.39216)(13101,0.57909)(13151,0.59658)(13201,0.60507)(13251,0.59580)(13301,0.60030)(13351,0.60019)(13401,0.59366)(13451,0.41370)(13501,0.57323)(13551,0.49217)(13601,0.57868)(13651,0.58466)(13701,0.56048)(13751,0.61299)(13801,0.57974)(13851,0.51119)(13901,0.59903)(13951,0.62856)(14001,0.56815)(14051,0.59421)(14101,0.59301)(14151,0.51264)(14201,0.61838)(14251,0.57202)(14301,0.56535)(14351,0.60996)(14401,0.51457)(14451,0.60933)(14501,0.53369)(14551,0.63260)(14601,0.61299)(14651,0.60277)(14701,0.54186)(14751,0.57335)(14801,0.60577)(14851,0.51515)(14901,0.50803)(14951,0.58209)(15001,0.60726)(15051,0.33360)(15101,0.52542)(15151,0.58934)(15201,0.62442)(15251,0.43548)(15301,0.57810)(15351,0.61308)(15401,0.59619)(15451,0.50565)(15501,0.58345)(15551,0.40722)(15601,0.60386)(15651,0.62672)(15701,0.60139)(15751,0.55631)(15801,0.55860)(15851,0.56105)(15901,0.60063)(15951,0.48225)(16001,0.61472)(16051,0.61615)(16101,0.60141)(16151,0.58299)(16201,0.62482)(16251,0.63821)
};
\addplot[ smooth,Set2-B,thick] plot coordinates {
(1,0.66751)(51,0.55061)(101,0.55478)(151,0.60815)(201,0.54714)(251,0.54856)(301,0.52946)(351,0.53901)(401,0.54911)(451,0.50959)(501,0.47305)(551,0.44862)(601,0.37135)(651,0.38556)(701,0.40996)(751,0.39905)(801,0.40867)(851,0.44834)(901,0.39332)(951,0.43134)(1001,0.44264)(1051,0.40927)(1101,0.41813)(1151,0.35410)(1201,0.36849)(1251,0.34443)(1301,0.36335)(1351,0.35052)(1401,0.33167)(1451,0.32332)(1501,0.32401)(1551,0.32287)(1601,0.35003)(1651,0.35398)(1701,0.33968)(1751,0.32405)(1801,0.36495)(1851,0.33260)(1901,0.31784)(1951,0.32702)(2001,0.33149)(2051,0.35159)(2101,0.32659)(2151,0.34627)(2201,0.35072)(2251,0.37538)(2301,0.32695)(2351,0.29403)(2401,0.34648)(2451,0.30194)(2501,0.34274)(2551,0.33281)(2601,0.34166)(2651,0.32512)(2701,0.35061)(2751,0.37549)(2801,0.28973)(2851,0.31573)(2901,0.33080)(2951,0.34723)(3001,0.31735)(3051,0.31457)(3101,0.31778)(3151,0.32798)(3201,0.32196)(3251,0.30801)(3301,0.36067)(3351,0.33125)(3401,0.33922)(3451,0.29635)(3501,0.32313)(3551,0.32569)(3601,0.32151)(3651,0.33968)(3701,0.33803)(3751,0.32186)(3801,0.34588)(3851,0.30475)(3901,0.34066)(3951,0.31276)(4001,0.37520)(4051,0.33162)(4101,0.31429)(4151,0.37568)(4201,0.32970)(4251,0.32656)(4301,0.36183)(4351,0.32376)(4401,0.31324)(4451,0.31099)(4501,0.34031)(4551,0.38142)(4601,0.34251)(4651,0.38004)(4701,0.36769)(4751,0.31433)(4801,0.33096)(4851,0.36684)(4901,0.32870)(4951,0.31787)(5001,0.30367)(5051,0.36360)(5101,0.33481)(5151,0.33980)(5201,0.32203)(5251,0.34568)(5301,0.32598)(5351,0.33917)(5401,0.39429)(5451,0.33672)(5501,0.31499)(5551,0.35280)(5601,0.37156)(5651,0.31871)(5701,0.34188)(5751,0.34256)(5801,0.36400)(5851,0.34832)(5901,0.36672)(5951,0.39437)(6001,0.31527)(6051,0.32257)(6101,0.38269)(6151,0.36422)(6201,0.37721)(6251,0.28266)(6301,0.38605)(6351,0.37909)(6401,0.32286)(6451,0.35787)(6501,0.28661)(6551,0.34939)(6601,0.36485)(6651,0.31678)(6701,0.39166)(6751,0.37650)(6801,0.39334)(6851,0.38652)(6901,0.37306)(6951,0.36444)(7001,0.35443)(7051,0.40031)(7101,0.39526)(7151,0.29892)(7201,0.37854)(7251,0.36846)(7301,0.37348)(7351,0.33250)(7401,0.37461)(7451,0.41060)(7501,0.35110)(7551,0.34518)(7601,0.39154)(7651,0.37434)(7701,0.41194)(7751,0.39222)(7801,0.38002)(7851,0.35083)(7901,0.36621)(7951,0.38873)(8001,0.30817)(8051,0.45917)(8101,0.36464)(8151,0.37512)(8201,0.35967)(8251,0.35159)(8301,0.42244)(8351,0.38483)(8401,0.35132)(8451,0.45134)(8501,0.43731)(8551,0.39622)(8601,0.36952)(8651,0.45142)(8701,0.43952)(8751,0.39845)(8801,0.41691)(8851,0.43111)(8901,0.45554)(8951,0.42158)(9001,0.46624)(9051,0.37048)(9101,0.32731)(9151,0.36707)(9201,0.43014)(9251,0.44410)(9301,0.43873)(9351,0.38753)(9401,0.39386)(9451,0.45318)(9501,0.43905)(9551,0.49825)(9601,0.46477)(9651,0.44261)(9701,0.52195)(9751,0.47576)(9801,0.44134)(9851,0.50983)(9901,0.47277)(9951,0.39807)(10001,0.46560)(10051,0.47527)(10101,0.47996)(10151,0.48132)(10201,0.26046)(10251,0.41080)(10301,0.48810)(10351,0.39417)(10401,0.46645)(10451,0.46165)(10501,0.40587)(10551,0.47648)(10601,0.36514)(10651,0.44848)(10701,0.47488)(10751,0.40421)(10801,0.40623)(10851,0.35914)(10901,0.45264)(10951,0.49205)(11001,0.51107)(11051,0.38958)(11101,0.49675)(11151,0.52708)(11201,0.36027)(11251,0.47543)(11301,0.52547)(11351,0.53749)(11401,0.45762)(11451,0.48074)(11501,0.43026)(11551,0.50479)(11601,0.49637)(11651,0.47954)(11701,0.50770)(11751,0.42394)(11801,0.52118)(11851,0.52461)(11901,0.49333)(11951,0.43385)(12001,0.51760)(12051,0.47586)(12101,0.52762)(12151,0.53970)(12201,0.51689)(12251,0.49184)(12301,0.42986)(12351,0.56529)(12401,0.55291)(12451,0.53235)(12501,0.40305)(12551,0.53399)(12601,0.53589)(12651,0.48309)(12701,0.47061)(12751,0.48583)(12801,0.52776)(12851,0.50215)(12901,0.49724)(12951,0.35254)(13001,0.52932)(13051,0.53174)(13101,0.45460)(13151,0.41044)(13201,0.52903)(13251,0.55511)(13301,0.51361)(13351,0.55897)(13401,0.54923)(13451,0.54761)(13501,0.51564)(13551,0.53506)(13601,0.52998)(13651,0.54049)(13701,0.54772)(13751,0.57640)(13801,0.44324)(13851,0.51313)(13901,0.55154)(13951,0.52858)(14001,0.53217)(14051,0.45328)(14101,0.54256)(14151,0.55845)(14201,0.52916)(14251,0.57535)(14301,0.57283)(14351,0.42032)(14401,0.55635)(14451,0.55879)(14501,0.49326)(14551,0.55367)(14601,0.56320)(14651,0.54509)(14701,0.51282)(14751,0.49079)(14801,0.57671)(14851,0.57439)(14901,0.57545)(14951,0.53714)(15001,0.48044)(15051,0.51631)(15101,0.57996)(15151,0.51839)(15201,0.50902)(15251,0.56065)(15301,0.48561)(15351,0.48619)(15401,0.54000)(15451,0.52526)(15501,0.25578)(15551,0.56024)(15601,0.58159)(15651,0.55948)(15701,0.55289)(15751,0.47617)(15801,0.55271)(15851,0.50234)(15901,0.52722)(15951,0.54151)(16001,0.57121)(16051,0.46976)(16101,0.53857)(16151,0.56767)(16201,0.58752)(16251,0.50229)};
\end{axis}
\end{tikzpicture}
}
\caption*{(c)}
\end{subfigure}
\begin{subfigure}[b]{0.24\textwidth}
\centering
\resizebox{\textwidth}{!}{
\begin{tikzpicture}
\begin{axis}[xlabel=\large Generator iteration,ylabel=\large $CE$ loss, xmin=-500,xmax=16000,ymin=-0.1,
ylabel near ticks,xlabel near ticks,
tick pos=left,tickwidth=1mm,legend pos= north east, legend entries={semi-supervised},axisStyle,legend style={font=\large}]
\addplot[ smooth,Set2-A,thick] plot coordinates {
(1,0.69160)(51,0.30992)(101,0.24731)(151,0.21015)(201,0.19277)(251,0.21383)(301,0.17599)(351,0.16477)(401,0.14163)(451,0.14978)(501,0.17117)(551,0.11814)(601,0.11449)(651,0.16944)(701,0.17665)(751,0.14754)(801,0.09043)(851,0.13720)(901,0.12087)(951,0.08486)(1001,0.12064)(1051,0.15174)(1101,0.09224)(1151,0.11567)(1201,0.14999)(1251,0.08876)(1301,0.13520)(1351,0.25517)(1401,0.10951)(1451,0.09742)(1501,0.08746)(1551,0.07899)(1601,0.05647)(1651,0.07363)(1701,0.03584)(1751,0.07746)(1801,0.07654)(1851,0.04906)(1901,0.10098)(1951,0.10125)(2001,0.05392)(2051,0.11177)(2101,0.08014)(2151,0.06228)(2201,0.05365)(2251,0.08554)(2301,0.08172)(2351,0.10149)(2401,0.08832)(2451,0.04686)(2501,0.03767)(2551,0.07508)(2601,0.05682)(2651,0.05890)(2701,0.05253)(2751,0.03741)(2801,0.07459)(2851,0.04700)(2901,0.04655)(2951,0.03930)(3001,0.07341)(3051,0.06374)(3101,0.03222)(3151,0.04128)(3201,0.04340)(3251,0.05268)(3301,0.05715)(3351,0.02594)(3401,0.05678)(3451,0.07827)(3501,0.05355)(3551,0.04272)(3601,0.04676)(3651,0.02684)(3701,0.03641)(3751,0.04493)(3801,0.03771)(3851,0.05776)(3901,0.06017)(3951,0.04603)(4001,0.07920)(4051,0.03634)(4101,0.02568)(4151,0.04062)(4201,0.03512)(4251,0.04959)(4301,0.03015)(4351,0.03268)(4401,0.05473)(4451,0.04741)(4501,0.04157)(4551,0.04704)(4601,0.03704)(4651,0.02403)(4701,0.03736)(4751,0.03177)(4801,0.04788)(4851,0.02950)(4901,0.03144)(4951,0.02174)(5001,0.02945)(5051,0.05707)(5101,0.02985)(5151,0.02784)(5201,0.02716)(5251,0.03991)(5301,0.01892)(5351,0.02283)(5401,0.03228)(5451,0.03042)(5501,0.03623)(5551,0.04934)(5601,0.02978)(5651,0.01695)(5701,0.04290)(5751,0.03494)(5801,0.04391)(5851,0.01572)(5901,0.02642)(5951,0.02867)(6001,0.03101)(6051,0.03643)(6101,0.03532)(6151,0.02383)(6201,0.03029)(6251,0.03525)(6301,0.02166)(6351,0.02255)(6401,0.03525)(6451,0.04223)(6501,0.01946)(6551,0.04543)(6601,0.01837)(6651,0.03231)(6701,0.02711)(6751,0.03426)(6801,0.02818)(6851,0.02449)(6901,0.03481)(6951,0.03864)(7001,0.02955)(7051,0.02133)(7101,0.03725)(7151,0.01259)(7201,0.06266)(7251,0.01746)(7301,0.04313)(7351,0.02638)(7401,0.01869)(7451,0.02032)(7501,0.04437)(7551,0.03529)(7601,0.04580)(7651,0.01545)(7701,0.03345)(7751,0.01401)(7801,0.02555)(7851,0.01470)(7901,0.03157)(7951,0.01406)(8001,0.02586)(8051,0.01261)(8101,0.02608)(8151,0.02962)(8201,0.02210)(8251,0.02463)(8301,0.01285)(8351,0.01719)(8401,0.03345)(8451,0.02497)(8501,0.01035)(8551,0.03100)(8601,0.02683)(8651,0.02959)(8701,0.01163)(8751,0.03408)(8801,0.02022)(8851,0.03349)(8901,0.02230)(8951,0.01523)(9001,0.01504)(9051,0.01324)(9101,0.04157)(9151,0.02593)(9201,0.01782)(9251,0.02217)(9301,0.02234)(9351,0.01588)(9401,0.02583)(9451,0.02515)(9501,0.02133)(9551,0.03756)(9601,0.02838)(9651,0.03099)(9701,0.01788)(9751,0.04424)(9801,0.01203)(9851,0.03096)(9901,0.01620)(9951,0.01921)(10001,0.03133)(10051,0.02736)(10101,0.04478)(10151,0.01973)(10201,0.02787)(10251,0.04353)(10301,0.00942)(10351,0.01022)(10401,0.01655)(10451,0.01021)(10501,0.02322)(10551,0.01548)(10601,0.05132)(10651,0.03397)(10701,0.03079)(10751,0.01670)(10801,0.01671)(10851,0.02191)(10901,0.01436)(10951,0.01595)(11001,0.05607)(11051,0.03412)(11101,0.02917)(11151,0.03596)(11201,0.02862)(11251,0.04780)(11301,0.01863)(11351,0.01004)(11401,0.01544)(11451,0.01611)(11501,0.01402)(11551,0.02347)(11601,0.01666)(11651,0.01518)(11701,0.01068)(11751,0.02295)(11801,0.01649)(11851,0.04096)(11901,0.04385)(11951,0.03957)(12001,0.02587)(12051,0.01346)(12101,0.02961)(12151,0.01962)(12201,0.01446)(12251,0.01429)(12301,0.04920)(12351,0.02058)(12401,0.01948)(12451,0.01928)(12501,0.00908)(12551,0.01000)(12601,0.01112)(12651,0.04010)(12701,0.00925)(12751,0.01760)(12801,0.01858)(12851,0.01931)(12901,0.01313)(12951,0.02736)(13001,0.01217)(13051,0.03731)(13101,0.02130)(13151,0.03029)(13201,0.03462)(13251,0.02386)(13301,0.00804)(13351,0.00976)(13401,0.02113)(13451,0.02153)(13501,0.01366)(13551,0.03838)(13601,0.01338)(13651,0.02497)(13701,0.01972)(13751,0.01903)(13801,0.01501)(13851,0.01832)(13901,0.00922)(13951,0.00736)(14001,0.00941)(14051,0.01813)(14101,0.01628)(14151,0.03782)(14201,0.01052)(14251,0.03339)(14301,0.01322)(14351,0.02324)(14401,0.01405)(14451,0.01253)(14501,0.00808)(14551,0.00888)(14601,0.01637)(14651,0.02463)(14701,0.03377)(14751,0.02286)(14801,0.01366)(14851,0.00407)(14901,0.03931)(14951,0.04060)(15001,0.01984)(15051,0.01250)(15101,0.02185)(15151,0.03404)(15201,0.01704)(15251,0.01408)(15301,0.02826)(15351,0.00849)(15401,0.00899)(15451,0.00738)(15501,0.00994)(15551,0.02161)(15601,0.01205)(15651,0.01957)(15701,0.01692)(15751,0.01107)(15801,0.00882)(15851,0.02788)(15901,0.02490)(15951,0.02315)(16001,0.00862)(16051,0.01734)(16101,0.00687)(16151,0.02733)(16201,0.01595)(16251,0.01251)
};
\end{axis}
\end{tikzpicture}
}
\caption*{(d)}
\end{subfigure}
\caption{Plot of the training loss during the unsupervised and semi-supervised training  of catWGAN. (a) is negative of the first two terms of $\mathcal{L}_{D_2}^{\text{WGAN}}$ (Equation~\ref{critic_loss}), representing the Wasserstein distance of the generated distribution and the real distribution. (b) is $S_r$ and (c) is $S_g$ as described in Section~\ref{catwgan} (Equation~\ref{e1}). (d) is the CE loss  for labeled samples. }
\label{train_stat}
\end{figure*}

\begin{figure}[tbp]
\centering
\includegraphics[width=0.5\textwidth]{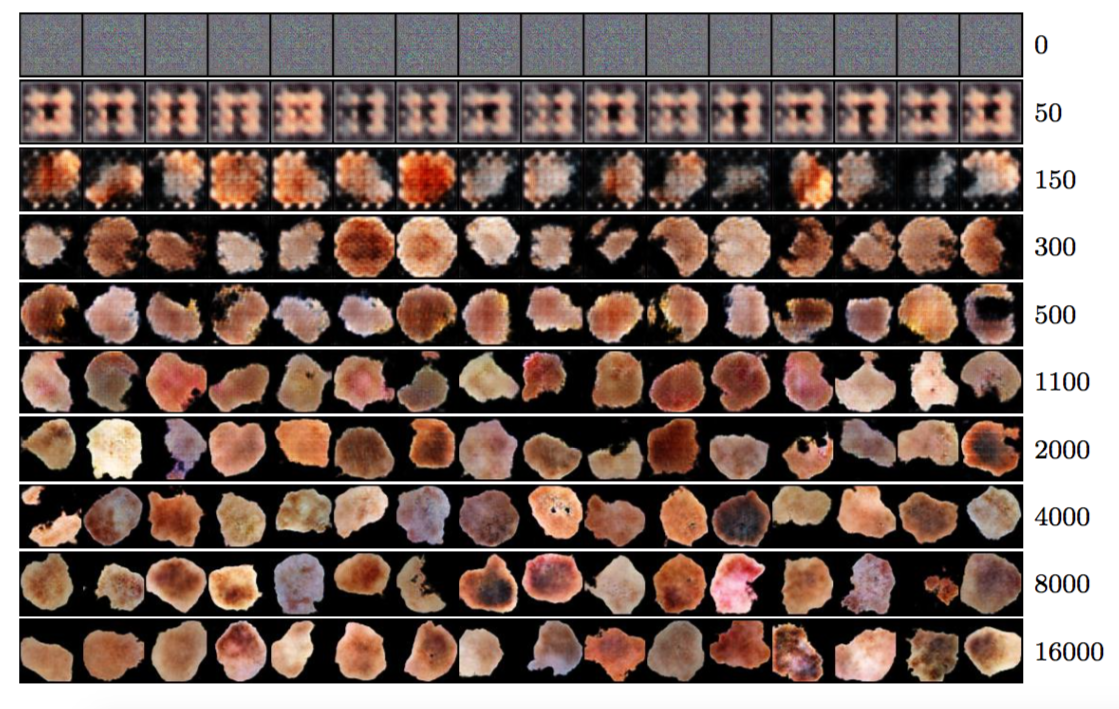}
\caption{ Generated images from the generator during the unsupervised training of catWGAN. Numbers shown on the right are the generator iteration number. 16 images are shown at each checkpoint. Each image is of size 64$\times$64. Semi-supervised catWAGAN's generated images (not shown here) exhibits similar trend.}
\label{evolve_g}
\end{figure}

\section{Results}
Figure~\ref{train_stat} shows the training loss statistics. (a) is the Wasserstein distance between the generated distribution and the real sample distribution and  was shown to conform to the generated image quality~\cite{arjovsky2017wasserstein} and we have found the same trend by examining the generated images at a series of checkpoints during the training as shown in  Figure~\ref{evolve_g}. (b) is the  estimated entropy of the predicted class distribution over real samples ($S_r$) and (c) is the estimated entropy of the predicted class distribution over generated samples ($S_g$). The loss in (b) increases in the first 3000 iterations and then gradually decreases. The loss in (c) displays an opposite trend. (d) shows a gradually decreasing  cross entropy loss for the labeled training samples in the semi-supervised learning.

\subsection{Evolution of  the feature representation during training}\label{evolve_feature}


The effectiveness of the features learnt in unsupervised setting was evaluated using cross-validation on the PH2 dataset. Linear SVM was used and the results reported here are from stratified 5-fold cross-validation. We fixed $c$ to be 1 in this experiment so that  the iteration number was the only varying parameter to be validated. Furthermore, since the PH2 dataset is an unbalanced dataset with  the amount of benign cases four times that of melanomas, the weights for melanoma samples were adjusted accordingly in the training.

As can be seen from Figure~\ref{val_loss}, for both unsupervised and semi-supervised training, the AC and AUC increases in the first 2000 iterations and then remains steady afterwards. The saturation of these two metrics is the result of large number of TN which makes small changes in TP less perceivable.  AP on the other hand also exhibits similar trend in the very beginning but decreases and saturates to a lower value.  On comparing Figure~\ref{val_loss} (b) with Figure~\ref{train_stat} (c),  we can see that, the best AP was achieved when $S_g$ is the smallest. Since the output of $\text{D}_1$ are two imaginary classes with no specific meaning,  $\text{D}_1$ could have learnt something trivial to separate the input into two classes (e.g. symmetric v.s. asymmertric), which do not well align with the desired separation (melanoma v.s. benign).

\begin{figure*}[htbp]
 \centering 
\begin{subfigure}[b]{0.3\textwidth}
\resizebox{\textwidth}{!}{
\begin{tikzpicture}
\begin{axis}[xlabel=\Large Generator iteration,ylabel=\Large AC, xmin=-500,xmax=16000, ymax=1,ymin=0,
ylabel near ticks,xlabel near ticks, legend pos= south east, legend entries={unsupervised, semi-supervised}, mark size=0.7pt,
tick pos=left,tickwidth=1mm,axisStyle,legend style={font=\large}]
\addplot[ smooth,Set2-A,thick] plot coordinates {
(0, 0.74)(50, 0.79)(100, 0.73)(150, 0.76)(200, 0.68)(250, 0.72)(300, 0.74)(350, 0.77)(400, 0.82)(450, 0.76)(500, 0.78)(550, 0.74)(600, 0.72)(650, 0.79)(700, 0.79)(750, 0.79)(800, 0.82)(850, 0.82)(900, 0.82)(950, 0.82)(1000, 0.83)(1050, 0.82)(1100, 0.83)(1150, 0.83)(1200, 0.82)(1250, 0.78)(1300, 0.79)(1350, 0.83)(1400, 0.79)(1450, 0.81)(1500, 0.77)(1550, 0.80)(1600, 0.77)(1650, 0.80)(1700, 0.78)(1750, 0.78)(1800, 0.84)(1850, 0.82)(1900, 0.84)(1950, 0.79)(2000, 0.82)(2050, 0.82)(2100, 0.81)(2150, 0.81)(2200, 0.83)(2250, 0.82)(2300, 0.82)(2350, 0.81)(2400, 0.84)(2450, 0.80)(2500, 0.78)(2550, 0.80)(2600, 0.82)(2650, 0.82)(2700, 0.82)(2750, 0.80)(2800, 0.82)(2850, 0.82)(2900, 0.83)(2950, 0.79)(3000, 0.81)(3050, 0.81)(3100, 0.84)(3150, 0.81)(3200, 0.83)(3250, 0.79)(3300, 0.77)(3350, 0.84)(3400, 0.80)(3450, 0.78)(3500, 0.83)(3550, 0.81)(3600, 0.82)(3650, 0.79)(3700, 0.77)(3750, 0.81)(3800, 0.82)(3850, 0.77)(3900, 0.83)(3950, 0.79)(4000, 0.76)(4050, 0.81)(4100, 0.82)(4150, 0.80)(4200, 0.79)(4250, 0.83)(4300, 0.77)(4350, 0.78)(4400, 0.79)(4450, 0.84)(4500, 0.81)(4550, 0.80)(4600, 0.79)(4650, 0.82)(4700, 0.83)(4750, 0.80)(4800, 0.80)(4850, 0.83)(4900, 0.81)(4950, 0.81)(5000, 0.81)(5050, 0.83)(5100, 0.83)(5150, 0.81)(5200, 0.82)(5250, 0.82)(5300, 0.83)(5350, 0.82)(5400, 0.82)(5450, 0.85)(5500, 0.83)(5550, 0.81)(5600, 0.82)(5650, 0.79)(5700, 0.80)(5750, 0.81)(5800, 0.82)(5850, 0.80)(5900, 0.80)(5950, 0.81)(6000, 0.82)(6050, 0.82)(6100, 0.79)(6150, 0.82)(6200, 0.81)(6250, 0.79)(6300, 0.80)(6350, 0.80)(6400, 0.82)(6450, 0.82)(6500, 0.80)(6550, 0.81)(6600, 0.82)(6650, 0.84)(6700, 0.80)(6750, 0.81)(6800, 0.82)(6850, 0.81)(6900, 0.80)(6950, 0.79)(7000, 0.85)(7050, 0.81)(7100, 0.82)(7150, 0.81)(7200, 0.86)(7250, 0.81)(7300, 0.81)(7350, 0.79)(7400, 0.81)(7450, 0.82)(7500, 0.85)(7550, 0.82)(7600, 0.81)(7650, 0.80)(7700, 0.82)(7750, 0.82)(7800, 0.83)(7850, 0.78)(7900, 0.78)(7950, 0.82)(8000, 0.81)(8050, 0.81)(8100, 0.82)(8150, 0.80)(8200, 0.82)(8250, 0.81)(8300, 0.81)(8350, 0.81)(8400, 0.84)(8450, 0.82)(8500, 0.79)(8550, 0.80)(8600, 0.80)(8650, 0.79)(8700, 0.80)(8750, 0.79)(8800, 0.80)(8850, 0.83)(8900, 0.82)(8950, 0.80)(9000, 0.81)(9050, 0.81)(9100, 0.83)(9150, 0.79)(9200, 0.81)(9250, 0.82)(9300, 0.78)(9350, 0.83)(9400, 0.81)(9450, 0.80)(9500, 0.81)(9550, 0.81)(9600, 0.79)(9650, 0.81)(9700, 0.80)(9750, 0.81)(9800, 0.80)(9850, 0.78)(9900, 0.79)(9950, 0.79)(10000, 0.84)(10050, 0.80)(10100, 0.78)(10150, 0.81)(10200, 0.80)(10250, 0.75)(10300, 0.81)(10350, 0.80)(10400, 0.82)(10450, 0.82)(10500, 0.80)(10550, 0.81)(10600, 0.76)(10650, 0.79)(10700, 0.79)(10750, 0.78)(10800, 0.81)(10850, 0.81)(10900, 0.81)(10950, 0.80)(11000, 0.80)(11050, 0.80)(11100, 0.82)(11150, 0.82)(11200, 0.81)(11250, 0.81)(11300, 0.80)(11350, 0.81)(11400, 0.82)(11450, 0.82)(11500, 0.80)(11550, 0.81)(11600, 0.79)(11650, 0.82)(11700, 0.82)(11750, 0.80)(11800, 0.83)(11850, 0.78)(11900, 0.83)(11950, 0.84)(12000, 0.80)(12050, 0.86)(12100, 0.82)(12150, 0.84)(12200, 0.82)(12250, 0.83)(12300, 0.84)(12350, 0.82)(12400, 0.83)(12450, 0.82)(12500, 0.81)(12550, 0.81)(12600, 0.82)(12650, 0.82)(12700, 0.81)(12750, 0.80)(12800, 0.82)(12850, 0.83)(12900, 0.81)(12950, 0.83)(13000, 0.81)(13050, 0.80)(13100, 0.80)(13150, 0.82)(13200, 0.82)(13250, 0.82)(13300, 0.83)(13350, 0.84)(13400, 0.81)(13450, 0.82)(13500, 0.82)(13550, 0.80)(13600, 0.78)(13650, 0.83)(13700, 0.80)(13750, 0.81)(13800, 0.80)(13850, 0.79)(13900, 0.83)(13950, 0.84)(14000, 0.77)(14050, 0.82)(14100, 0.82)(14150, 0.81)(14200, 0.82)(14250, 0.83)(14300, 0.82)(14350, 0.81)(14400, 0.83)(14450, 0.84)(14500, 0.84)(14550, 0.81)(14600, 0.82)(14650, 0.82)(14700, 0.82)(14750, 0.81)(14800, 0.83)(14850, 0.83)(14900, 0.78)(14950, 0.83)(15000, 0.83)(15050, 0.83)(15100, 0.84)(15150, 0.80)(15200, 0.82)(15250, 0.80)(15300, 0.81)(15350, 0.79)(15400, 0.81)(15450, 0.81)(15500, 0.83)(15550, 0.83)(15600, 0.82)(15650, 0.82)(15700, 0.82)(15750, 0.80)(15800, 0.80)(15850, 0.81)(15900, 0.80)(15950, 0.81)(16000, 0.81)(16050, 0.80)(16100, 0.81)(16150, 0.80)(16200, 0.83)
};
\addplot[ smooth,Set2-B] plot coordinates {
(0, 0.74)(50, 0.83)(100, 0.78)(150, 0.80)(200, 0.83)(250, 0.83)(300, 0.80)(350, 0.80)(400, 0.80)(450, 0.81)(500, 0.81)(550, 0.80)(600, 0.82)(650, 0.86)(700, 0.85)(750, 0.86)(800, 0.86)(850, 0.86)(900, 0.87)(950, 0.85)(1000, 0.83)(1050, 0.86)(1100, 0.84)(1150, 0.85)(1200, 0.86)(1250, 0.85)(1300, 0.86)(1350, 0.85)(1400, 0.86)(1450, 0.87)(1500, 0.86)(1550, 0.85)(1600, 0.87)(1650, 0.86)(1700, 0.85)(1750, 0.84)(1800, 0.85)(1850, 0.84)(1900, 0.83)(1950, 0.84)(2000, 0.88)(2050, 0.83)(2100, 0.83)(2150, 0.83)(2200, 0.85)(2250, 0.86)(2300, 0.83)(2350, 0.83)(2400, 0.82)(2450, 0.83)(2500, 0.83)(2550, 0.84)(2600, 0.84)(2650, 0.85)(2700, 0.83)(2750, 0.85)(2800, 0.84)(2850, 0.84)(2900, 0.85)(2950, 0.84)(3000, 0.82)(3050, 0.84)(3100, 0.84)(3150, 0.82)(3200, 0.84)(3250, 0.83)(3300, 0.83)(3350, 0.83)(3400, 0.84)(3450, 0.81)(3500, 0.83)(3550, 0.82)(3600, 0.83)(3650, 0.83)(3700, 0.84)(3750, 0.83)(3800, 0.84)(3850, 0.86)(3900, 0.84)(3950, 0.86)(4000, 0.85)(4050, 0.84)(4100, 0.82)(4150, 0.85)(4200, 0.86)(4250, 0.86)(4300, 0.88)(4350, 0.85)(4400, 0.84)(4450, 0.87)(4500, 0.85)(4550, 0.86)(4600, 0.88)(4650, 0.87)(4700, 0.86)(4750, 0.87)(4800, 0.85)(4850, 0.87)(4900, 0.86)(4950, 0.84)(5000, 0.86)(5050, 0.85)(5100, 0.85)(5150, 0.86)(5200, 0.86)(5250, 0.86)(5300, 0.86)(5350, 0.83)(5400, 0.86)(5450, 0.84)(5500, 0.85)(5550, 0.87)(5600, 0.86)(5650, 0.88)(5700, 0.85)(5750, 0.88)(5800, 0.88)(5850, 0.88)(5900, 0.86)(5950, 0.87)(6000, 0.89)(6050, 0.88)(6100, 0.86)(6150, 0.89)(6200, 0.88)(6250, 0.86)(6300, 0.86)(6350, 0.88)(6400, 0.86)(6450, 0.88)(6500, 0.88)(6550, 0.86)(6600, 0.85)(6650, 0.86)(6700, 0.89)(6750, 0.88)(6800, 0.89)(6850, 0.89)(6900, 0.88)(6950, 0.88)(7000, 0.86)(7050, 0.88)(7100, 0.89)(7150, 0.89)(7200, 0.91)(7250, 0.91)(7300, 0.89)(7350, 0.88)(7400, 0.90)(7450, 0.88)(7500, 0.87)(7550, 0.88)(7600, 0.89)(7650, 0.89)(7700, 0.88)(7750, 0.90)(7800, 0.88)(7850, 0.89)(7900, 0.89)(7950, 0.87)(8000, 0.88)(8050, 0.88)(8100, 0.88)(8150, 0.86)(8200, 0.89)(8250, 0.88)(8300, 0.88)(8350, 0.86)(8400, 0.89)(8450, 0.89)(8500, 0.85)(8550, 0.86)(8600, 0.87)(8650, 0.85)(8700, 0.86)(8750, 0.86)(8800, 0.88)(8850, 0.88)(8900, 0.88)(8950, 0.87)(9000, 0.88)(9050, 0.88)(9100, 0.88)(9150, 0.84)(9200, 0.88)(9250, 0.86)(9300, 0.83)(9350, 0.89)(9400, 0.84)(9450, 0.88)(9500, 0.88)(9550, 0.86)(9600, 0.89)(9650, 0.87)(9700, 0.86)(9750, 0.89)(9800, 0.86)(9850, 0.89)(9900, 0.88)(9950, 0.86)(10000, 0.86)(10050, 0.87)(10100, 0.88)(10150, 0.88)(10200, 0.88)(10250, 0.87)(10300, 0.86)(10350, 0.86)(10400, 0.86)(10450, 0.87)(10500, 0.88)(10550, 0.85)(10600, 0.86)(10650, 0.86)(10700, 0.86)(10750, 0.88)(10800, 0.87)(10850, 0.87)(10900, 0.88)(10950, 0.86)(11000, 0.89)(11050, 0.89)(11100, 0.84)(11150, 0.89)(11200, 0.86)(11250, 0.86)(11300, 0.85)(11350, 0.86)(11400, 0.85)(11450, 0.87)(11500, 0.86)(11550, 0.86)(11600, 0.88)(11650, 0.87)(11700, 0.87)(11750, 0.87)(11800, 0.86)(11850, 0.85)(11900, 0.86)(11950, 0.89)(12000, 0.85)(12050, 0.86)(12100, 0.87)(12150, 0.83)(12200, 0.83)(12250, 0.83)(12300, 0.86)(12350, 0.88)(12400, 0.84)(12450, 0.87)(12500, 0.86)(12550, 0.86)(12600, 0.86)(12650, 0.86)(12700, 0.83)(12750, 0.86)(12800, 0.89)(12850, 0.87)(12900, 0.88)(12950, 0.87)(13000, 0.88)(13050, 0.88)(13100, 0.86)(13150, 0.88)(13200, 0.86)(13250, 0.85)(13300, 0.86)(13350, 0.84)(13400, 0.86)(13450, 0.87)(13500, 0.87)(13550, 0.86)(13600, 0.87)(13650, 0.86)(13700, 0.84)(13750, 0.86)(13800, 0.87)(13850, 0.86)(13900, 0.86)(13950, 0.86)(14000, 0.85)(14050, 0.86)(14100, 0.85)(14150, 0.88)(14200, 0.83)(14250, 0.86)(14300, 0.86)(14350, 0.87)(14400, 0.84)(14450, 0.86)(14500, 0.86)(14550, 0.86)(14600, 0.87)(14650, 0.86)(14700, 0.86)(14750, 0.86)(14800, 0.85)(14850, 0.86)(14900, 0.88)(14950, 0.88)(15000, 0.86)(15050, 0.86)(15100, 0.86)(15150, 0.84)(15200, 0.86)(15250, 0.88)(15300, 0.85)(15350, 0.86)(15400, 0.88)(15450, 0.86)(15500, 0.87)(15550, 0.85)(15600, 0.86)(15650, 0.86)(15700, 0.87)(15750, 0.87)(15800, 0.84)(15850, 0.88)(15900, 0.85)(15950, 0.86)(16000, 0.86)(16050, 0.87)(16100, 0.86)(16150, 0.86)(16200, 0.87)(16250, 0.88)(16300, 0.88)(16350, 0.88)(16400, 0.85)(16450, 0.88)(16500, 0.88)(16550, 0.89)(16600, 0.86)(16650, 0.87)(16700, 0.85)(16750, 0.86)
};
\end{axis}
\end{tikzpicture}
}
\caption*{(a)}
\end{subfigure}
\begin{subfigure}[b]{0.3\textwidth}
\resizebox{\textwidth}{!}{
\begin{tikzpicture}
\begin{axis}[xlabel=\Large Generator iteration,ylabel=\Large AP, xmin=-500,xmax=16000, ymax=1,ymin=0,
ylabel near ticks,xlabel near ticks,legend pos= south east,
tick pos=left,tickwidth=1mm, legend entries={unsupervised, semi-supervised, semi-supervised },axisStyle,legend style={font=\large}]
\addplot[ smooth,Set2-A,thick] plot coordinates {
(0, 0.68)(50, 0.76)(100, 0.62)(150, 0.62)(200, 0.59)(250, 0.56)(300, 0.64)(350, 0.74)(400, 0.74)(450, 0.65)(500, 0.62)(550, 0.60)(600, 0.54)(650, 0.62)(700, 0.65)(750, 0.66)(800, 0.67)(850, 0.67)(900, 0.75)(950, 0.72)(1000, 0.76)(1050, 0.72)(1100, 0.75)(1150, 0.75)(1200, 0.79)(1250, 0.74)(1300, 0.76)(1350, 0.77)(1400, 0.76)(1450, 0.75)(1500, 0.73)(1550, 0.73)(1600, 0.68)(1650, 0.77)(1700, 0.69)(1750, 0.67)(1800, 0.81)(1850, 0.76)(1900, 0.75)(1950, 0.69)(2000, 0.78)(2050, 0.74)(2100, 0.79)(2150, 0.61)(2200, 0.72)(2250, 0.66)(2300, 0.65)(2350, 0.65)(2400, 0.71)(2450, 0.71)(2500, 0.67)(2550, 0.70)(2600, 0.70)(2650, 0.69)(2700, 0.66)(2750, 0.70)(2800, 0.67)(2850, 0.70)(2900, 0.68)(2950, 0.68)(3000, 0.68)(3050, 0.67)(3100, 0.72)(3150, 0.68)(3200, 0.73)(3250, 0.62)(3300, 0.66)(3350, 0.71)(3400, 0.62)(3450, 0.63)(3500, 0.66)(3550, 0.63)(3600, 0.68)(3650, 0.63)(3700, 0.60)(3750, 0.65)(3800, 0.68)(3850, 0.66)(3900, 0.72)(3950, 0.66)(4000, 0.68)(4050, 0.67)(4100, 0.67)(4150, 0.66)(4200, 0.67)(4250, 0.68)(4300, 0.66)(4350, 0.68)(4400, 0.68)(4450, 0.73)(4500, 0.68)(4550, 0.66)(4600, 0.67)(4650, 0.69)(4700, 0.66)(4750, 0.70)(4800, 0.65)(4850, 0.66)(4900, 0.67)(4950, 0.70)(5000, 0.69)(5050, 0.68)(5100, 0.69)(5150, 0.67)(5200, 0.67)(5250, 0.67)(5300, 0.68)(5350, 0.67)(5400, 0.68)(5450, 0.71)(5500, 0.71)(5550, 0.72)(5600, 0.72)(5650, 0.68)(5700, 0.70)(5750, 0.70)(5800, 0.66)(5850, 0.70)(5900, 0.71)(5950, 0.75)(6000, 0.71)(6050, 0.69)(6100, 0.70)(6150, 0.69)(6200, 0.71)(6250, 0.69)(6300, 0.68)(6350, 0.70)(6400, 0.67)(6450, 0.68)(6500, 0.67)(6550, 0.69)(6600, 0.70)(6650, 0.73)(6700, 0.73)(6750, 0.73)(6800, 0.69)(6850, 0.72)(6900, 0.67)(6950, 0.70)(7000, 0.74)(7050, 0.70)(7100, 0.70)(7150, 0.72)(7200, 0.74)(7250, 0.71)(7300, 0.72)(7350, 0.67)(7400, 0.70)(7450, 0.68)(7500, 0.70)(7550, 0.67)(7600, 0.70)(7650, 0.68)(7700, 0.74)(7750, 0.70)(7800, 0.73)(7850, 0.69)(7900, 0.67)(7950, 0.75)(8000, 0.73)(8050, 0.72)(8100, 0.71)(8150, 0.69)(8200, 0.70)(8250, 0.67)(8300, 0.68)(8350, 0.69)(8400, 0.73)(8450, 0.71)(8500, 0.68)(8550, 0.69)(8600, 0.71)(8650, 0.64)(8700, 0.71)(8750, 0.67)(8800, 0.65)(8850, 0.69)(8900, 0.72)(8950, 0.67)(9000, 0.65)(9050, 0.69)(9100, 0.68)(9150, 0.66)(9200, 0.73)(9250, 0.71)(9300, 0.67)(9350, 0.74)(9400, 0.67)(9450, 0.70)(9500, 0.72)(9550, 0.68)(9600, 0.71)(9650, 0.67)(9700, 0.64)(9750, 0.70)(9800, 0.67)(9850, 0.67)(9900, 0.67)(9950, 0.65)(10000, 0.70)(10050, 0.71)(10100, 0.66)(10150, 0.69)(10200, 0.65)(10250, 0.69)(10300, 0.69)(10350, 0.70)(10400, 0.70)(10450, 0.70)(10500, 0.68)(10550, 0.70)(10600, 0.66)(10650, 0.69)(10700, 0.66)(10750, 0.64)(10800, 0.70)(10850, 0.70)(10900, 0.71)(10950, 0.70)(11000, 0.66)(11050, 0.67)(11100, 0.70)(11150, 0.69)(11200, 0.70)(11250, 0.66)(11300, 0.67)(11350, 0.70)(11400, 0.70)(11450, 0.71)(11500, 0.71)(11550, 0.68)(11600, 0.72)(11650, 0.69)(11700, 0.71)(11750, 0.70)(11800, 0.77)(11850, 0.69)(11900, 0.70)(11950, 0.73)(12000, 0.71)(12050, 0.75)(12100, 0.74)(12150, 0.77)(12200, 0.72)(12250, 0.73)(12300, 0.74)(12350, 0.72)(12400, 0.69)(12450, 0.71)(12500, 0.69)(12550, 0.72)(12600, 0.69)(12650, 0.75)(12700, 0.72)(12750, 0.67)(12800, 0.70)(12850, 0.73)(12900, 0.71)(12950, 0.68)(13000, 0.68)(13050, 0.69)(13100, 0.70)(13150, 0.73)(13200, 0.72)(13250, 0.70)(13300, 0.73)(13350, 0.71)(13400, 0.70)(13450, 0.74)(13500, 0.72)(13550, 0.68)(13600, 0.69)(13650, 0.71)(13700, 0.71)(13750, 0.71)(13800, 0.71)(13850, 0.66)(13900, 0.69)(13950, 0.71)(14000, 0.69)(14050, 0.68)(14100, 0.67)(14150, 0.68)(14200, 0.69)(14250, 0.68)(14300, 0.65)(14350, 0.67)(14400, 0.71)(14450, 0.69)(14500, 0.71)(14550, 0.69)(14600, 0.69)(14650, 0.68)(14700, 0.68)(14750, 0.67)(14800, 0.70)(14850, 0.72)(14900, 0.70)(14950, 0.69)(15000, 0.72)(15050, 0.68)(15100, 0.69)(15150, 0.69)(15200, 0.68)(15250, 0.68)(15300, 0.69)(15350, 0.65)(15400, 0.66)(15450, 0.66)(15500, 0.70)(15550, 0.68)(15600, 0.70)(15650, 0.67)(15700, 0.68)(15750, 0.68)(15800, 0.66)(15850, 0.69)(15900, 0.68)(15950, 0.72)(16000, 0.67)(16050, 0.66)(16100, 0.65)(16150, 0.68)(16200, 0.70)
};
\addplot[ smooth,Set2-B] plot coordinates {
(0, 0.59)(50, 0.67)(100, 0.70)(150, 0.68)(200, 0.75)(250, 0.79)(300, 0.71)(350, 0.70)(400, 0.74)(450, 0.77)(500, 0.76)(550, 0.74)(600, 0.76)(650, 0.79)(700, 0.82)(750, 0.83)(800, 0.83)(850, 0.83)(900, 0.85)(950, 0.86)(1000, 0.83)(1050, 0.85)(1100, 0.81)(1150, 0.83)(1200, 0.81)(1250, 0.82)(1300, 0.82)(1350, 0.84)(1400, 0.83)(1450, 0.84)(1500, 0.83)(1550, 0.85)(1600, 0.82)(1650, 0.82)(1700, 0.83)(1750, 0.82)(1800, 0.82)(1850, 0.82)(1900, 0.80)(1950, 0.83)(2000, 0.83)(2050, 0.83)(2100, 0.82)(2150, 0.82)(2200, 0.82)(2250, 0.82)(2300, 0.80)(2350, 0.78)(2400, 0.80)(2450, 0.80)(2500, 0.78)(2550, 0.80)(2600, 0.81)(2650, 0.79)(2700, 0.79)(2750, 0.80)(2800, 0.79)(2850, 0.81)(2900, 0.80)(2950, 0.79)(3000, 0.77)(3050, 0.82)(3100, 0.77)(3150, 0.78)(3200, 0.79)(3250, 0.79)(3300, 0.77)(3350, 0.76)(3400, 0.77)(3450, 0.76)(3500, 0.80)(3550, 0.77)(3600, 0.77)(3650, 0.77)(3700, 0.78)(3750, 0.77)(3800, 0.77)(3850, 0.80)(3900, 0.75)(3950, 0.78)(4000, 0.79)(4050, 0.78)(4100, 0.76)(4150, 0.77)(4200, 0.77)(4250, 0.79)(4300, 0.79)(4350, 0.78)(4400, 0.80)(4450, 0.79)(4500, 0.78)(4550, 0.77)(4600, 0.79)(4650, 0.79)(4700, 0.81)(4750, 0.81)(4800, 0.79)(4850, 0.81)(4900, 0.79)(4950, 0.78)(5000, 0.79)(5050, 0.80)(5100, 0.80)(5150, 0.80)(5200, 0.80)(5250, 0.81)(5300, 0.81)(5350, 0.80)(5400, 0.81)(5450, 0.78)(5500, 0.78)(5550, 0.81)(5600, 0.79)(5650, 0.79)(5700, 0.80)(5750, 0.80)(5800, 0.81)(5850, 0.80)(5900, 0.82)(5950, 0.83)(6000, 0.80)(6050, 0.83)(6100, 0.83)(6150, 0.83)(6200, 0.83)(6250, 0.83)(6300, 0.78)(6350, 0.84)(6400, 0.81)(6450, 0.84)(6500, 0.83)(6550, 0.81)(6600, 0.83)(6650, 0.83)(6700, 0.83)(6750, 0.84)(6800, 0.83)(6850, 0.84)(6900, 0.83)(6950, 0.84)(7000, 0.81)(7050, 0.82)(7100, 0.82)(7150, 0.82)(7200, 0.83)(7250, 0.84)(7300, 0.84)(7350, 0.83)(7400, 0.84)(7450, 0.84)(7500, 0.81)(7550, 0.84)(7600, 0.81)(7650, 0.82)(7700, 0.79)(7750, 0.82)(7800, 0.81)(7850, 0.82)(7900, 0.83)(7950, 0.78)(8000, 0.80)(8050, 0.81)(8100, 0.81)(8150, 0.81)(8200, 0.82)(8250, 0.82)(8300, 0.82)(8350, 0.81)(8400, 0.84)(8450, 0.81)(8500, 0.82)(8550, 0.80)(8600, 0.79)(8650, 0.79)(8700, 0.81)(8750, 0.80)(8800, 0.82)(8850, 0.79)(8900, 0.80)(8950, 0.80)(9000, 0.79)(9050, 0.79)(9100, 0.82)(9150, 0.80)(9200, 0.79)(9250, 0.82)(9300, 0.81)(9350, 0.82)(9400, 0.78)(9450, 0.83)(9500, 0.81)(9550, 0.79)(9600, 0.79)(9650, 0.81)(9700, 0.81)(9750, 0.83)(9800, 0.80)(9850, 0.82)(9900, 0.79)(9950, 0.81)(10000, 0.83)(10050, 0.82)(10100, 0.83)(10150, 0.83)(10200, 0.81)(10250, 0.81)(10300, 0.82)(10350, 0.83)(10400, 0.80)(10450, 0.79)(10500, 0.78)(10550, 0.77)(10600, 0.80)(10650, 0.80)(10700, 0.81)(10750, 0.79)(10800, 0.79)(10850, 0.82)(10900, 0.80)(10950, 0.83)(11000, 0.80)(11050, 0.82)(11100, 0.81)(11150, 0.81)(11200, 0.78)(11250, 0.81)(11300, 0.79)(11350, 0.80)(11400, 0.79)(11450, 0.82)(11500, 0.79)(11550, 0.82)(11600, 0.81)(11650, 0.82)(11700, 0.80)(11750, 0.84)(11800, 0.81)(11850, 0.80)(11900, 0.80)(11950, 0.81)(12000, 0.82)(12050, 0.81)(12100, 0.81)(12150, 0.80)(12200, 0.78)(12250, 0.78)(12300, 0.78)(12350, 0.82)(12400, 0.78)(12450, 0.80)(12500, 0.78)(12550, 0.82)(12600, 0.79)(12650, 0.78)(12700, 0.78)(12750, 0.80)(12800, 0.80)(12850, 0.79)(12900, 0.81)(12950, 0.78)(13000, 0.80)(13050, 0.79)(13100, 0.78)(13150, 0.77)(13200, 0.78)(13250, 0.80)(13300, 0.82)(13350, 0.79)(13400, 0.78)(13450, 0.81)(13500, 0.81)(13550, 0.80)(13600, 0.77)(13650, 0.77)(13700, 0.78)(13750, 0.80)(13800, 0.81)(13850, 0.79)(13900, 0.77)(13950, 0.79)(14000, 0.79)(14050, 0.79)(14100, 0.80)(14150, 0.79)(14200, 0.79)(14250, 0.78)(14300, 0.81)(14350, 0.80)(14400, 0.75)(14450, 0.80)(14500, 0.80)(14550, 0.79)(14600, 0.78)(14650, 0.78)(14700, 0.79)(14750, 0.79)(14800, 0.79)(14850, 0.81)(14900, 0.81)(14950, 0.83)(15000, 0.82)(15050, 0.78)(15100, 0.79)(15150, 0.77)(15200, 0.78)(15250, 0.81)(15300, 0.77)(15350, 0.80)(15400, 0.82)(15450, 0.79)(15500, 0.80)(15550, 0.78)(15600, 0.76)(15650, 0.79)(15700, 0.81)(15750, 0.78)(15800, 0.76)(15850, 0.81)(15900, 0.76)(15950, 0.77)(16000, 0.82)(16050, 0.82)(16100, 0.77)(16150, 0.82)(16200, 0.81)(16250, 0.84)(16300, 0.76)(16350, 0.82)(16400, 0.81)(16450, 0.81)(16500, 0.81)(16550, 0.81)(16600, 0.79)(16650, 0.80)(16700, 0.79)(16750, 0.77)};
\end{axis}
\end{tikzpicture}
}
\caption*{(b)}
\end{subfigure}
\begin{subfigure}[b]{0.3\textwidth}
\resizebox{\textwidth}{!}{
\begin{tikzpicture}
\begin{axis}[xlabel=\Large Generator iteration,ylabel=\Large AUC, xmin=-500,xmax=16000, ymax=1,ymin=0,
ylabel near ticks,xlabel near ticks,legend pos= south east, legend entries={unsupervised, semi-supervised},
tick pos=left,tickwidth=1mm,axisStyle,legend style={font=\large}]
\addplot[ smooth,Set2-A,thick] plot coordinates {
(0, 0.82)(50, 0.90)(100, 0.82)(150, 0.82)(200, 0.81)(250, 0.82)(300, 0.84)(350, 0.88)(400, 0.88)(450, 0.84)(500, 0.84)(550, 0.79)(600, 0.76)(650, 0.79)(700, 0.81)(750, 0.83)(800, 0.84)(850, 0.82)(900, 0.88)(950, 0.86)(1000, 0.88)(1050, 0.85)(1100, 0.87)(1150, 0.88)(1200, 0.89)(1250, 0.87)(1300, 0.89)(1350, 0.89)(1400, 0.88)(1450, 0.91)(1500, 0.88)(1550, 0.88)(1600, 0.85)(1650, 0.90)(1700, 0.84)(1750, 0.85)(1800, 0.90)(1850, 0.88)(1900, 0.89)(1950, 0.87)(2000, 0.89)(2050, 0.88)(2100, 0.89)(2150, 0.84)(2200, 0.89)(2250, 0.85)(2300, 0.86)(2350, 0.84)(2400, 0.88)(2450, 0.87)(2500, 0.87)(2550, 0.86)(2600, 0.87)(2650, 0.87)(2700, 0.85)(2750, 0.87)(2800, 0.86)(2850, 0.88)(2900, 0.86)(2950, 0.85)(3000, 0.86)(3050, 0.87)(3100, 0.88)(3150, 0.85)(3200, 0.89)(3250, 0.86)(3300, 0.83)(3350, 0.89)(3400, 0.83)(3450, 0.86)(3500, 0.85)(3550, 0.84)(3600, 0.86)(3650, 0.84)(3700, 0.83)(3750, 0.86)(3800, 0.86)(3850, 0.84)(3900, 0.89)(3950, 0.85)(4000, 0.87)(4050, 0.87)(4100, 0.87)(4150, 0.87)(4200, 0.87)(4250, 0.87)(4300, 0.84)(4350, 0.86)(4400, 0.87)(4450, 0.88)(4500, 0.88)(4550, 0.86)(4600, 0.86)(4650, 0.88)(4700, 0.86)(4750, 0.87)(4800, 0.84)(4850, 0.87)(4900, 0.87)(4950, 0.88)(5000, 0.87)(5050, 0.88)(5100, 0.87)(5150, 0.87)(5200, 0.87)(5250, 0.86)(5300, 0.88)(5350, 0.87)(5400, 0.86)(5450, 0.89)(5500, 0.88)(5550, 0.87)(5600, 0.88)(5650, 0.87)(5700, 0.87)(5750, 0.87)(5800, 0.85)(5850, 0.88)(5900, 0.87)(5950, 0.88)(6000, 0.86)(6050, 0.86)(6100, 0.87)(6150, 0.86)(6200, 0.86)(6250, 0.87)(6300, 0.86)(6350, 0.86)(6400, 0.85)(6450, 0.86)(6500, 0.84)(6550, 0.86)(6600, 0.86)(6650, 0.88)(6700, 0.89)(6750, 0.88)(6800, 0.86)(6850, 0.88)(6900, 0.84)(6950, 0.87)(7000, 0.90)(7050, 0.86)(7100, 0.87)(7150, 0.88)(7200, 0.89)(7250, 0.88)(7300, 0.87)(7350, 0.85)(7400, 0.87)(7450, 0.86)(7500, 0.86)(7550, 0.85)(7600, 0.86)(7650, 0.87)(7700, 0.88)(7750, 0.88)(7800, 0.88)(7850, 0.85)(7900, 0.84)(7950, 0.89)(8000, 0.88)(8050, 0.89)(8100, 0.87)(8150, 0.86)(8200, 0.86)(8250, 0.85)(8300, 0.86)(8350, 0.87)(8400, 0.88)(8450, 0.87)(8500, 0.86)(8550, 0.87)(8600, 0.87)(8650, 0.84)(8700, 0.87)(8750, 0.86)(8800, 0.83)(8850, 0.88)(8900, 0.88)(8950, 0.84)(9000, 0.83)(9050, 0.86)(9100, 0.86)(9150, 0.85)(9200, 0.87)(9250, 0.87)(9300, 0.84)(9350, 0.89)(9400, 0.86)(9450, 0.85)(9500, 0.86)(9550, 0.86)(9600, 0.88)(9650, 0.85)(9700, 0.84)(9750, 0.86)(9800, 0.86)(9850, 0.84)(9900, 0.86)(9950, 0.83)(10000, 0.87)(10050, 0.86)(10100, 0.84)(10150, 0.85)(10200, 0.84)(10250, 0.85)(10300, 0.86)(10350, 0.87)(10400, 0.86)(10450, 0.87)(10500, 0.85)(10550, 0.87)(10600, 0.84)(10650, 0.87)(10700, 0.84)(10750, 0.86)(10800, 0.87)(10850, 0.87)(10900, 0.88)(10950, 0.86)(11000, 0.87)(11050, 0.87)(11100, 0.87)(11150, 0.85)(11200, 0.87)(11250, 0.85)(11300, 0.85)(11350, 0.87)(11400, 0.86)(11450, 0.87)(11500, 0.86)(11550, 0.86)(11600, 0.87)(11650, 0.86)(11700, 0.88)(11750, 0.86)(11800, 0.90)(11850, 0.86)(11900, 0.88)(11950, 0.89)(12000, 0.86)(12050, 0.88)(12100, 0.89)(12150, 0.90)(12200, 0.86)(12250, 0.88)(12300, 0.89)(12350, 0.88)(12400, 0.86)(12450, 0.86)(12500, 0.85)(12550, 0.87)(12600, 0.85)(12650, 0.89)(12700, 0.88)(12750, 0.85)(12800, 0.87)(12850, 0.87)(12900, 0.87)(12950, 0.84)(13000, 0.86)(13050, 0.85)(13100, 0.86)(13150, 0.87)(13200, 0.87)(13250, 0.86)(13300, 0.88)(13350, 0.85)(13400, 0.85)(13450, 0.88)(13500, 0.86)(13550, 0.86)(13600, 0.86)(13650, 0.86)(13700, 0.86)(13750, 0.86)(13800, 0.86)(13850, 0.84)(13900, 0.85)(13950, 0.87)(14000, 0.85)(14050, 0.86)(14100, 0.85)(14150, 0.85)(14200, 0.86)(14250, 0.85)(14300, 0.83)(14350, 0.85)(14400, 0.88)(14450, 0.86)(14500, 0.87)(14550, 0.85)(14600, 0.84)(14650, 0.85)(14700, 0.86)(14750, 0.85)(14800, 0.86)(14850, 0.87)(14900, 0.86)(14950, 0.86)(15000, 0.87)(15050, 0.85)(15100, 0.86)(15150, 0.86)(15200, 0.86)(15250, 0.85)(15300, 0.86)(15350, 0.84)(15400, 0.83)(15450, 0.84)(15500, 0.87)(15550, 0.85)(15600, 0.87)(15650, 0.84)(15700, 0.85)(15750, 0.85)(15800, 0.84)(15850, 0.84)(15900, 0.85)(15950, 0.87)(16000, 0.85)(16050, 0.84)(16100, 0.85)(16150, 0.85)(16200, 0.87)
};
\addplot[ smooth,Set2-B] plot coordinates {
(0, 0.79)(50, 0.83)(100, 0.84)(150, 0.85)(200, 0.89)(250, 0.91)(300, 0.89)(350, 0.89)(400, 0.89)(450, 0.89)(500, 0.89)(550, 0.87)(600, 0.87)(650, 0.90)(700, 0.91)(750, 0.92)(800, 0.91)(850, 0.92)(900, 0.93)(950, 0.93)(1000, 0.92)(1050, 0.93)(1100, 0.91)(1150, 0.92)(1200, 0.90)(1250, 0.91)(1300, 0.91)(1350, 0.92)(1400, 0.92)(1450, 0.92)(1500, 0.93)(1550, 0.93)(1600, 0.92)(1650, 0.92)(1700, 0.93)(1750, 0.92)(1800, 0.91)(1850, 0.92)(1900, 0.91)(1950, 0.91)(2000, 0.92)(2050, 0.92)(2100, 0.92)(2150, 0.92)(2200, 0.91)(2250, 0.92)(2300, 0.91)(2350, 0.90)(2400, 0.91)(2450, 0.90)(2500, 0.90)(2550, 0.91)(2600, 0.90)(2650, 0.91)(2700, 0.91)(2750, 0.91)(2800, 0.90)(2850, 0.91)(2900, 0.90)(2950, 0.89)(3000, 0.89)(3050, 0.91)(3100, 0.89)(3150, 0.89)(3200, 0.90)(3250, 0.91)(3300, 0.89)(3350, 0.89)(3400, 0.89)(3450, 0.89)(3500, 0.90)(3550, 0.89)(3600, 0.89)(3650, 0.89)(3700, 0.89)(3750, 0.89)(3800, 0.88)(3850, 0.90)(3900, 0.89)(3950, 0.90)(4000, 0.90)(4050, 0.90)(4100, 0.88)(4150, 0.90)(4200, 0.89)(4250, 0.91)(4300, 0.90)(4350, 0.89)(4400, 0.90)(4450, 0.90)(4500, 0.90)(4550, 0.89)(4600, 0.90)(4650, 0.90)(4700, 0.90)(4750, 0.91)(4800, 0.90)(4850, 0.91)(4900, 0.90)(4950, 0.89)(5000, 0.89)(5050, 0.89)(5100, 0.90)(5150, 0.91)(5200, 0.90)(5250, 0.90)(5300, 0.90)(5350, 0.90)(5400, 0.91)(5450, 0.88)(5500, 0.88)(5550, 0.90)(5600, 0.90)(5650, 0.89)(5700, 0.89)(5750, 0.89)(5800, 0.90)(5850, 0.91)(5900, 0.90)(5950, 0.91)(6000, 0.91)(6050, 0.91)(6100, 0.91)(6150, 0.91)(6200, 0.91)(6250, 0.91)(6300, 0.87)(6350, 0.91)(6400, 0.90)(6450, 0.92)(6500, 0.91)(6550, 0.90)(6600, 0.91)(6650, 0.92)(6700, 0.91)(6750, 0.92)(6800, 0.92)(6850, 0.92)(6900, 0.90)(6950, 0.92)(7000, 0.90)(7050, 0.91)(7100, 0.90)(7150, 0.92)(7200, 0.90)(7250, 0.93)(7300, 0.92)(7350, 0.91)(7400, 0.91)(7450, 0.92)(7500, 0.90)(7550, 0.93)(7600, 0.91)(7650, 0.92)(7700, 0.90)(7750, 0.90)(7800, 0.92)(7850, 0.91)(7900, 0.91)(7950, 0.89)(8000, 0.91)(8050, 0.91)(8100, 0.89)(8150, 0.91)(8200, 0.90)(8250, 0.90)(8300, 0.92)(8350, 0.91)(8400, 0.91)(8450, 0.91)(8500, 0.91)(8550, 0.90)(8600, 0.89)(8650, 0.89)(8700, 0.90)(8750, 0.91)(8800, 0.90)(8850, 0.88)(8900, 0.90)(8950, 0.90)(9000, 0.87)(9050, 0.88)(9100, 0.90)(9150, 0.88)(9200, 0.88)(9250, 0.91)(9300, 0.91)(9350, 0.90)(9400, 0.88)(9450, 0.91)(9500, 0.90)(9550, 0.88)(9600, 0.88)(9650, 0.89)(9700, 0.89)(9750, 0.90)(9800, 0.89)(9850, 0.90)(9900, 0.90)(9950, 0.91)(10000, 0.91)(10050, 0.91)(10100, 0.91)(10150, 0.89)(10200, 0.89)(10250, 0.89)(10300, 0.89)(10350, 0.90)(10400, 0.89)(10450, 0.89)(10500, 0.89)(10550, 0.88)(10600, 0.90)(10650, 0.89)(10700, 0.90)(10750, 0.90)(10800, 0.89)(10850, 0.90)(10900, 0.90)(10950, 0.90)(11000, 0.89)(11050, 0.90)(11100, 0.89)(11150, 0.89)(11200, 0.87)(11250, 0.90)(11300, 0.89)(11350, 0.89)(11400, 0.90)(11450, 0.90)(11500, 0.88)(11550, 0.89)(11600, 0.88)(11650, 0.90)(11700, 0.89)(11750, 0.91)(11800, 0.90)(11850, 0.89)(11900, 0.89)(11950, 0.90)(12000, 0.89)(12050, 0.90)(12100, 0.90)(12150, 0.90)(12200, 0.89)(12250, 0.89)(12300, 0.89)(12350, 0.90)(12400, 0.88)(12450, 0.90)(12500, 0.89)(12550, 0.91)(12600, 0.89)(12650, 0.90)(12700, 0.89)(12750, 0.90)(12800, 0.90)(12850, 0.89)(12900, 0.91)(12950, 0.90)(13000, 0.88)(13050, 0.88)(13100, 0.88)(13150, 0.89)(13200, 0.89)(13250, 0.89)(13300, 0.90)(13350, 0.89)(13400, 0.89)(13450, 0.90)(13500, 0.90)(13550, 0.90)(13600, 0.88)(13650, 0.89)(13700, 0.88)(13750, 0.89)(13800, 0.90)(13850, 0.89)(13900, 0.88)(13950, 0.88)(14000, 0.87)(14050, 0.89)(14100, 0.89)(14150, 0.88)(14200, 0.89)(14250, 0.89)(14300, 0.89)(14350, 0.91)(14400, 0.88)(14450, 0.89)(14500, 0.89)(14550, 0.89)(14600, 0.89)(14650, 0.88)(14700, 0.88)(14750, 0.89)(14800, 0.89)(14850, 0.91)(14900, 0.90)(14950, 0.91)(15000, 0.90)(15050, 0.89)(15100, 0.89)(15150, 0.88)(15200, 0.89)(15250, 0.90)(15300, 0.87)(15350, 0.89)(15400, 0.89)(15450, 0.89)(15500, 0.89)(15550, 0.89)(15600, 0.87)(15650, 0.88)(15700, 0.91)(15750, 0.88)(15800, 0.87)(15850, 0.90)(15900, 0.88)(15950, 0.89)(16000, 0.92)(16050, 0.92)(16100, 0.88)(16150, 0.91)(16200, 0.90)(16250, 0.92)(16300, 0.89)(16350, 0.91)(16400, 0.89)(16450, 0.90)(16500, 0.90)(16550, 0.91)(16600, 0.89)(16650, 0.90)(16700, 0.90)(16750, 0.86)};
\end{axis}
\end{tikzpicture}
}
\caption*{(c)}
\end{subfigure}
\caption{Performance of features learned by catWGAN in the process of training validated on PH2 dataset. From left to right shows the AC, AP, AUC respectively. Green line shows the result of unsupervised learning whereas orange line shows semi-supervised learning with 140 labeled images. }
\label{val_loss}
\end{figure*}
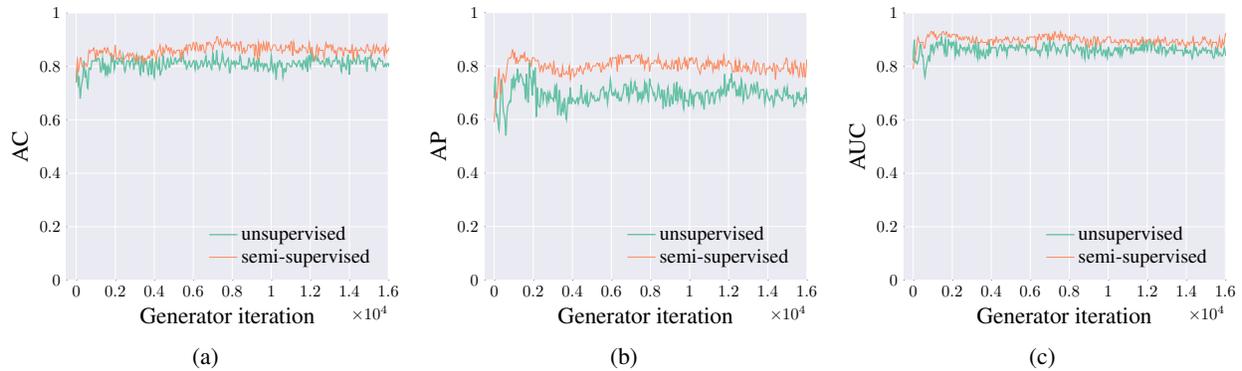

\begin{figure}[htbp]
 \centering 
\begin{tikzpicture}
\begin{axis}[xlabel= False positive rate (1-SP),ylabel=True positive rate (SE), xmin=0,xmax=1, ymax=1,ymin=0,
ylabel near ticks,xlabel near ticks, legend pos= south east, legend entries={catWGAN-unsuper, catWGAN-semi, DAE}, mark size=0.7pt,
tick pos=left,tickwidth=0mm,
axis background/.style={fill=axisbk},
    grid=both,
    grid style={line width=.1pt, draw=white},
    legend style={fill=none, draw=none},
    legend cell align=left,
    axis line style={draw=none}]
\addplot[ smooth,Set2-A,thick] plot coordinates {
(0.00, 0.01)(0.00, 0.07)(0.01, 0.07)(0.01, 0.08)(0.01, 0.08)(0.01, 0.12)(0.02, 0.12)(0.02, 0.13)(0.05, 0.13)(0.05, 0.15)(0.07, 0.15)(0.07, 0.16)(0.07, 0.16)(0.07, 0.20)(0.09, 0.20)(0.09, 0.21)(0.09, 0.21)(0.09, 0.23)(0.10, 0.23)(0.10, 0.24)(0.11, 0.24)(0.11, 0.25)(0.12, 0.25)(0.12, 0.28)(0.12, 0.28)(0.12, 0.31)(0.13, 0.31)(0.13, 0.32)(0.15, 0.32)(0.15, 0.33)(0.15, 0.33)(0.15, 0.35)(0.19, 0.35)(0.19, 0.36)(0.20, 0.36)(0.20, 0.37)(0.24, 0.37)(0.24, 0.39)(0.25, 0.39)(0.25, 0.40)(0.26, 0.40)(0.26, 0.44)(0.27, 0.44)(0.27, 0.45)(0.27, 0.45)(0.27, 0.47)(0.28, 0.47)(0.28, 0.48)(0.30, 0.48)(0.30, 0.49)(0.31, 0.49)(0.31, 0.51)(0.32, 0.51)(0.32, 0.52)(0.34, 0.52)(0.34, 0.53)(0.36, 0.53)(0.36, 0.55)(0.37, 0.55)(0.37, 0.56)(0.37, 0.56)(0.37, 0.57)(0.38, 0.57)(0.38, 0.59)(0.43, 0.59)(0.43, 0.60)(0.50, 0.60)(0.50, 0.61)(0.51, 0.61)(0.51, 0.63)(0.51, 0.63)(0.51, 0.64)(0.51, 0.64)(0.51, 0.65)(0.53, 0.65)(0.53, 0.67)(0.54, 0.67)(0.54, 0.68)(0.59, 0.68)(0.59, 0.69)(0.61, 0.69)(0.61, 0.72)(0.62, 0.72)(0.62, 0.73)(0.67, 0.73)(0.67, 0.75)(0.68, 0.75)(0.68, 0.76)(0.71, 0.76)(0.71, 0.77)(0.72, 0.77)(0.72, 0.79)(0.76, 0.79)(0.76, 0.81)(0.77, 0.81)(0.77, 0.83)(0.78, 0.83)(0.78, 0.85)(0.79, 0.85)(0.79, 0.87)(0.79, 0.87)(0.79, 0.88)(0.82, 0.88)(0.82, 0.89)(0.85, 0.89)(0.85, 0.91)(0.85, 0.91)(0.85, 0.92)(0.87, 0.92)(0.87, 0.93)(0.88, 0.93)(0.88, 0.95)(0.89, 0.95)(0.89, 0.96)(0.90, 0.96)(0.90, 0.97)(0.96, 0.97)(0.96, 1.00)(1.00, 1.00)
};
\addplot[ smooth,Set2-B,thick] plot coordinates {
(0.00, 0.01)(0.00, 0.07)(0.01, 0.07)(0.01, 0.11)(0.01, 0.11)(0.01, 0.12)(0.01, 0.12)(0.01, 0.13)(0.03, 0.13)(0.03, 0.15)(0.03, 0.15)(0.03, 0.16)(0.04, 0.16)(0.04, 0.19)(0.04, 0.19)(0.04, 0.21)(0.05, 0.21)(0.05, 0.23)(0.05, 0.23)(0.05, 0.24)(0.06, 0.24)(0.06, 0.25)(0.10, 0.25)(0.10, 0.27)(0.11, 0.27)(0.11, 0.31)(0.12, 0.31)(0.12, 0.32)(0.13, 0.32)(0.13, 0.35)(0.13, 0.35)(0.13, 0.36)(0.14, 0.36)(0.14, 0.37)(0.14, 0.37)(0.14, 0.39)(0.15, 0.39)(0.15, 0.44)(0.15, 0.44)(0.15, 0.48)(0.16, 0.48)(0.16, 0.49)(0.17, 0.49)(0.17, 0.51)(0.21, 0.51)(0.21, 0.52)(0.21, 0.52)(0.21, 0.53)(0.22, 0.53)(0.22, 0.55)(0.22, 0.55)(0.22, 0.56)(0.24, 0.56)(0.24, 0.57)(0.25, 0.57)(0.25, 0.59)(0.27, 0.59)(0.27, 0.60)(0.30, 0.60)(0.30, 0.61)(0.30, 0.61)(0.30, 0.63)(0.31, 0.63)(0.31, 0.64)(0.34, 0.64)(0.34, 0.65)(0.34, 0.65)(0.34, 0.67)(0.38, 0.67)(0.38, 0.68)(0.42, 0.68)(0.42, 0.69)(0.43, 0.69)(0.43, 0.71)(0.44, 0.71)(0.44, 0.72)(0.49, 0.72)(0.49, 0.73)(0.51, 0.73)(0.51, 0.75)(0.54, 0.75)(0.54, 0.76)(0.56, 0.76)(0.56, 0.77)(0.59, 0.77)(0.59, 0.81)(0.60, 0.81)(0.60, 0.83)(0.66, 0.83)(0.66, 0.84)(0.68, 0.84)(0.68, 0.85)(0.69, 0.85)(0.69, 0.87)(0.72, 0.87)(0.72, 0.88)(0.75, 0.88)(0.75, 0.89)(0.86, 0.89)(0.86, 0.92)(0.88, 0.92)(0.88, 0.93)(0.93, 0.93)(0.93, 0.95)(0.94, 0.95)(0.94, 0.96)(0.96, 0.96)(0.96, 0.97)(0.97, 0.97)(0.97, 0.99)(0.98, 0.99)(0.98, 1.00)(1.00, 1.00)
};
\addplot[ smooth,Set2-C,thick] plot coordinates {
(0.00, 0.01)(0.01, 0.01)(0.01, 0.04)(0.01, 0.04)(0.01, 0.05)(0.03, 0.05)(0.03, 0.11)(0.03, 0.11)(0.03, 0.13)(0.04, 0.13)(0.04, 0.15)(0.04, 0.15)(0.04, 0.16)(0.04, 0.16)(0.04, 0.17)(0.05, 0.17)(0.05, 0.20)(0.06, 0.20)(0.06, 0.23)(0.07, 0.23)(0.07, 0.24)(0.07, 0.24)(0.07, 0.27)(0.10, 0.27)(0.10, 0.28)(0.11, 0.28)(0.11, 0.29)(0.11, 0.29)(0.11, 0.31)(0.12, 0.31)(0.12, 0.32)(0.13, 0.32)(0.13, 0.33)(0.13, 0.33)(0.13, 0.35)(0.15, 0.35)(0.15, 0.36)(0.17, 0.36)(0.17, 0.37)(0.21, 0.37)(0.21, 0.39)(0.22, 0.39)(0.22, 0.40)(0.22, 0.40)(0.22, 0.41)(0.23, 0.41)(0.23, 0.43)(0.27, 0.43)(0.27, 0.44)(0.27, 0.44)(0.27, 0.45)(0.29, 0.45)(0.29, 0.47)(0.30, 0.47)(0.30, 0.48)(0.30, 0.48)(0.30, 0.49)(0.31, 0.49)(0.31, 0.51)(0.34, 0.51)(0.34, 0.52)(0.38, 0.52)(0.38, 0.53)(0.40, 0.53)(0.40, 0.55)(0.42, 0.55)(0.42, 0.56)(0.44, 0.56)(0.44, 0.60)(0.46, 0.60)(0.46, 0.61)(0.48, 0.61)(0.48, 0.63)(0.52, 0.63)(0.52, 0.64)(0.52, 0.64)(0.52, 0.65)(0.53, 0.65)(0.53, 0.68)(0.53, 0.68)(0.53, 0.69)(0.53, 0.69)(0.53, 0.71)(0.54, 0.71)(0.54, 0.72)(0.54, 0.72)(0.54, 0.73)(0.57, 0.73)(0.57, 0.75)(0.60, 0.75)(0.60, 0.76)(0.61, 0.76)(0.61, 0.77)(0.62, 0.77)(0.62, 0.79)(0.66, 0.79)(0.66, 0.80)(0.67, 0.80)(0.67, 0.83)(0.71, 0.83)(0.71, 0.84)(0.75, 0.84)(0.75, 0.87)(0.77, 0.87)(0.77, 0.89)(0.78, 0.89)(0.78, 0.91)(0.78, 0.91)(0.78, 0.93)(0.81, 0.93)(0.81, 0.95)(0.86, 0.95)(0.86, 0.96)(0.90, 0.96)(0.90, 0.97)(0.95, 0.97)(0.95, 0.99)(0.98, 0.99)(0.98, 1.00)(1.00, 1.00)
};
\draw[gray,dashed](axis cs:0,0) -- (axis cs:1,1);

\end{axis}
\end{tikzpicture}

\caption{ROC curves of the best proposed and baseline models on the 2016 ISIC challenge test dataset.}
\label{roc}
\end{figure}
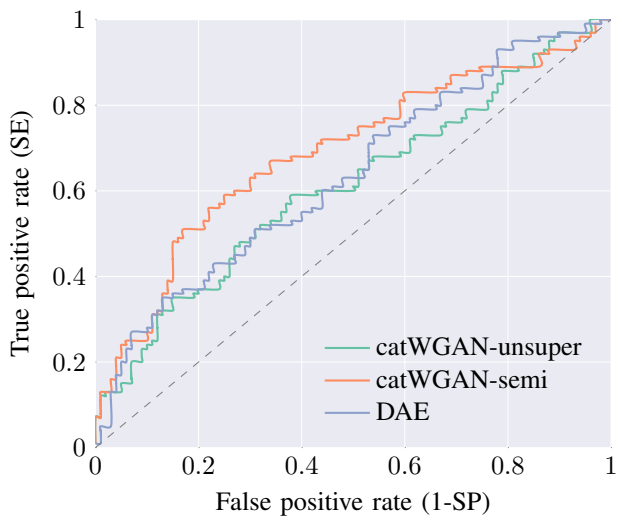

\subsection{Comparison with baselines}\label{compare_baseline}
With the validation results on PH2 dataset from Figure~\ref{val_loss}, we selected an ensemble of models  that gives the best AP scores (around 2000 iterations) for the final testing on the 2016 ISIC test set.    Results from different models are averaged to give the final result. Similar scheme was performed for DAE.   Test results are shown in Table~\ref{t1}. Unsupervised catWGAN performs slightly better than DAE in terms of AP and the performance of both methods is in between that of the edge and color histogram. Semi-supervised  catWGAN achieves much better results than these four methods.

Figure~\ref{roc} demonstrates the  receiver operating characteristic (ROC) curve for both the unsupervised and semi-supervised catWGAN and the baseline DAE on the 2016 ISIC test set. We can see performance improvements at almost all operating points for the semi-supervised method over the unsupervised method.

\begin{table}[tb]
\tiny
\centering
\begin{tabular}{cccccc} \toprule
Method				&	AP		&	AUC		&	AC				\\ \midrule
Edge Histogram		&	0.265	&	0.571	&	 0.665		\\  
Color Histogram		&	0.36	 	&	0.626	&	 0.789			\\ 
DAE					&	0.329	&	0.634	&	0.794			\\
catWGAN-unsuper (proposed)				&	0.351 	&	0.613	&	 0.812		\\
catWGAN-semi (proposed)				&	0.424 	&	0.690	&	 0.81			\\   \bottomrule \vspace{8pt}
\end{tabular}
\caption{Comparison to the baseline methods on the test set of the 2016 ISIC challenge dataset.}
\label{t1}
\end{table}%

\begin{figure}[tb]
\centering
\includegraphics[width=0.5\textwidth]{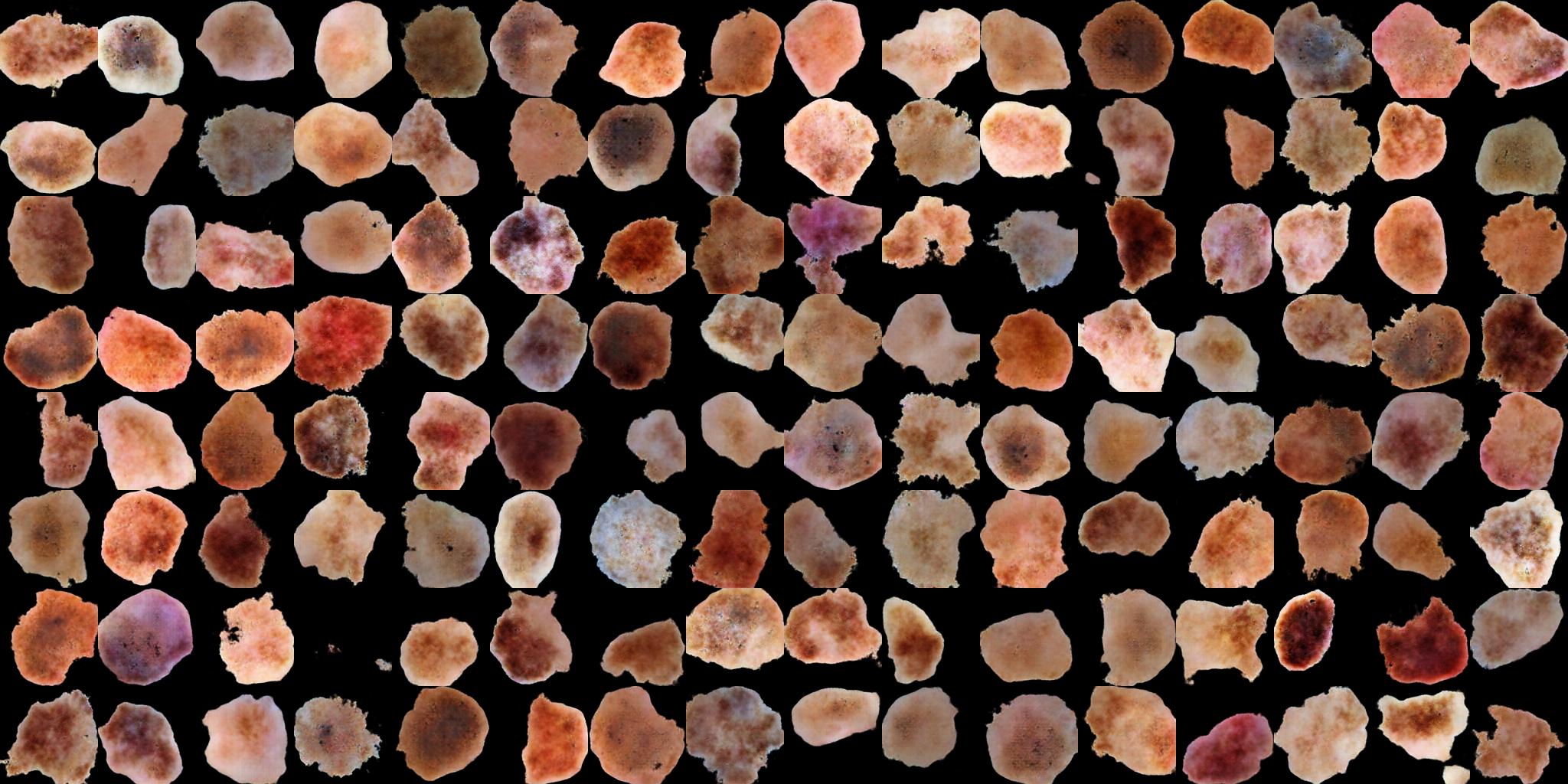}
\caption{Generated dermoscopy images of size $128\times128$ by using the proposed unsupervised method.}
\label{large}
\end{figure}

\section{Discussion}

From the unsupervised results of the first experiment, we can observe that even if the generator continues to improve the generated image, the features learnt from discriminator  could oscillate instead of improving. It implicitly shows that without supervision, the network could not learn task dependent features, even if the objective of $\text{D}_1$ was to produce confidence values of two classes (expected to be melanoma and benign). Comparing the classification result of the unsupervised feature to the results of the edge and color histogram, we can see that both unsupervised catWGAN and DAE have their AP stuck in between the two simple hand-crafted features. This evidence further suggests the network might have learnt   to separate the samples into two trivial classes, such as red vs orange, or symmetric vs asymmetric. Therefore, for images with variations in both color, shape and texture, unsupervised training might have limitations in capturing the desired task specific features. However, for images with distinct structural variations, like digits, unsupervised training will still be useful.


In this work, we have focused on the binary classification problem, but this method can be easily extended to multiple classes  by replacing the last two layers of $\text{D}_1$ with the desired number of classes. Even in cases where the dataset contains images of only one semantic class, the output of $\text{D}_1$ can still be arbitrary number of imaginary classes, with each one  corresponding to  potential attributes. 

There are some limitations of this work that we want to address. First, the generated images are only of a spatial size of $64\times 64$ which makes the classification result (AP 0.424) not directly comparable to the state-of-the-art supervised melanoma classification methods (AP 0.624~\cite{yu2017automated}). We have applied our proposed method on larger images of size $128\times 128$. Good looking images are generated as shown in Figure~\ref{large} but the features are not so robust. Effective architecture should be explored for $\text{D}_1$ in better learning of the feature representation in larger spatial resolution. Second, to make our results comparable with that of the other methods working on the 2016 ISIC challenge dataset, we restricted ourself on the 900 training images in the evaluation. Despite the fact that the size of unlabeled dataset used for training is augmented to 20k,  the effective distinct number of images could limit the performance of the proposed method. In the future, we would like to use the full set 13,000 images of ISIC Archive as the unlabeled dataset and all the 900 labeled images in the semi-supervised setting to evaluate the full potential. Third, we believe explicit coordination between the supervised and unsupervised task should be explored in the future to ensure a more robust feature learning.

\section{Conclusion}
In this work, we used categorial generative adversarial network assisted by Wasserstein distance for both unsupervised and semi-supervised learning. By just using 70 labeled samples from each class, the proposed model is able to learn a feature representation whose performance is much better than the denoising autoencoder and simple hand-crafted features.   This demonstrates the efficacy of the proposed method and its application  in many other medical image classification problems where feature representation is desired but limited labeled data is available. Another advantage of our proposed method is the ability to generate real-world like dermoscopy images with various shape, colour and surface texture.

\section{Acknowledgement}
This research was enabled in part by support provided by \footnote{http://www.calculquebec.ca/en/ }{Calcul Qu\'ebec}.


\bibliographystyle{spbasic} 

\bibliography{dermo,gan}

\end{document}